\documentclass[10pt,twocolumn,letterpaper]{article}

\usepackage[pagenumbers]{cvpr} 

\usepackage{multirow}
\usepackage{multicol}
\usepackage{array}
\usepackage{colortbl}
\usepackage{tabularx}
\usepackage{makecell}
\usepackage{cases}

\usepackage[dvipsnames]{xcolor}

\definecolor{cvprblue}{rgb}{0.21,0.49,0.74}
\usepackage[pagebackref,breaklinks,colorlinks,citecolor=cvprblue]{hyperref}

\usepackage{amssymb}
\usepackage{pifont}
\newcommand{\xmark}{\ding{55}}%

\newcolumntype{Y}{>{\centering\arraybackslash}X}
\newcolumntype{Z}{>{\hsize=.75\hsize\linewidth=\hsize\centering\arraybackslash}X}
\newcolumntype{V}{>{\hsize=1.2\hsize\linewidth=\hsize\centering\arraybackslash}X}

\newlength\paramarginsize 
\newcommand{\paramargin}{\vspace{\paramarginsize}}
\newcommand{\topic}[1]{\paramargin\noindent \textbf{#1}\ }

\def\paperID{4758} 
\def\confName{CVPR}
\def\confYear{2024}

\title{Beyond Entropy: \\Style Transfer Guided Single Image Continual Test-Time Adaptation}

\author{Younggeol Cho\footnotemark[1]
\qquad Youngrae Kim\footnotemark[1]
\qquad Dongman Lee \\
School of Computing, Korea Advanced Institute of Science and Technology (KAIST)\\
Daejeon, Republic of Korea\\
{\tt\small \{rangewing, youngrae.kim, dlee\}@kaist.ac.kr}
}

\definecolor{lightgray}{gray}{0.9}

\begin{document}
\maketitle
\begin{abstract}
Continual test-time adaptation (cTTA) methods are designed to facilitate the continual adaptation of models to dynamically changing real-world environments where computational resources are limited. 
Due to this inherent limitation, existing approaches fail to simultaneously achieve accuracy and efficiency.
In detail, when using a single image, the instability caused by batch normalization layers and entropy loss significantly destabilizes many existing methods in real-world cTTA scenarios.
To overcome these challenges, we present BESTTA, a novel single image continual test-time adaptation method guided by style transfer, which enables stable and efficient adaptation to the target environment by transferring the style of the input image to the source style.
To implement the proposed method, we devise BeIN, a simple yet powerful normalization method, along with the style-guided losses.
We demonstrate that BESTTA effectively adapts to the continually changing target environment, leveraging only a single image on both semantic segmentation and image classification tasks. 
Remarkably, despite training only two parameters in a BeIN layer consuming the least memory, BESTTA outperforms existing state-of-the-art methods in terms of performance. Our code is available at (TBD).

\end{abstract}    
\footnotetext[1]{\ Equal contribution}
\section{Introduction}
\label{sec:intro}

\begin{figure}[ht]
    \centering
    \includegraphics[width=\linewidth]{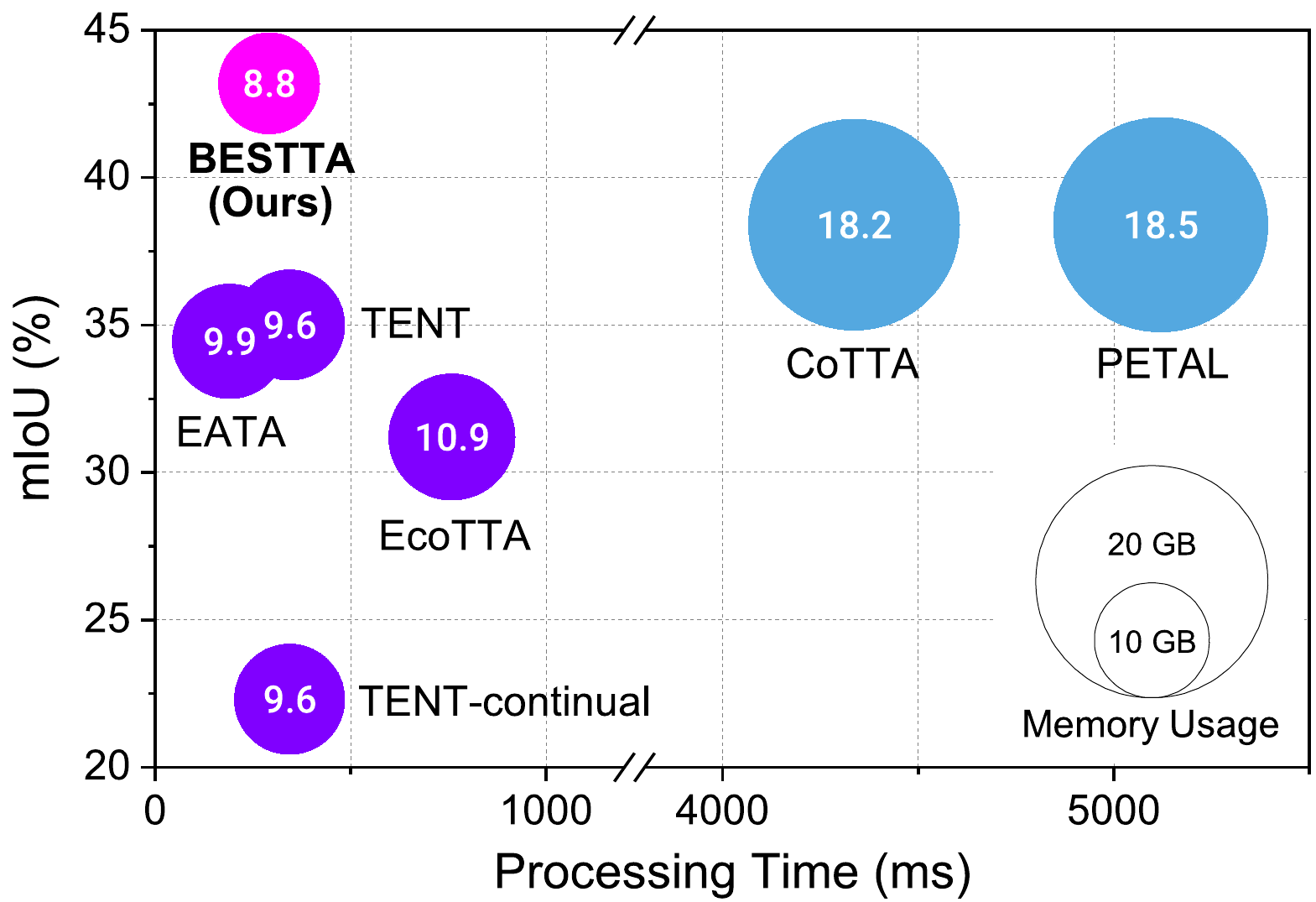}
    \caption{Semantic segmentation performances on the Cityscapes-to-ACDC single image continual test-time adaptation task, evaluated on DeepLabV3Plus-ResNet50~\cite{chen2018deeplabv3+}. Processing time includes inference and adaptation time per image. 
    The circle radius and values indicate the peak GPU memory usage. The violet circles are batch normalization-based and the blue circles are pseudo-label-based methods. BESTTA significantly outperforms the state-of-the-art continual test-time adaptation methods~\cite{wang2021tent,song2023ecotta,niu2022efficient,wang2022continual,Brahma2023probabilistic} in terms of mIoU while consuming the least GPU memory. }
    \label{fig:teaser}
\end{figure}

Deep learning models have significantly improved the performance of computer vision applications~\cite{he2016deep,dosovitskiy2020image}. However, in real-world scenarios, the models suffer from performance degradation due to domain shifts between training and target data. Unsupervised domain adaptation (UDA)~\cite{pan2010domain_uda,sun2019unsupervised_uda,patel2015visual_uda,long2016unsupervised_uda,tsai2018learning_uda_adv,ganin2015unsupervised_uda_adv,long2015learning_uda_discrepancy_loss}
and test-time training (TTT)~\cite{sun2020test_ttt, liu2021ttt++}
techniques have been proposed to address the performance gap. 
These methods assume that the models are adapted to the unseen domain with the huge amount of source data at test time. However it is impractical to utilize these methods in the real world due to the limited computational resources.
Fully test-time adaptation (TTA) methods~\cite{wang2021tent,niu2022towards_SAR,chen2022contrastivetta} have recently emerged to enable online adaptation of a trained model to the target environment without source data or labels. 
These TTA methods have limitations in that they can only adapt to a single target domain, resulting in overfitting to the target domain and forgetting the prebuilt knowledge.
Consequently, recent studies~\cite{wang2022continual,niu2022efficient,song2023ecotta,Brahma2023probabilistic} have proposed continual test-time adaptation (cTTA) methods to adapt a model for continually changing target domains over a long period, addressing the issues of catastrophic forgetting and error accumulation found in the previous TTA methods.

\begin{table*}[t!]
\centering
\def\arraystretch{1}
{\small
\setlength{\tabcolsep}{7pt}
\begin{tabular}{lcccccc}
\hline
\toprule
\multirow{2}{*}{Settings} & \multicolumn{2}{c}{Data} & \multicolumn{2}{c}{Learning}  & \multirow{2}{*}{\shortstack{Continually \\ Changing Domain }} & \multirow{2}{*}{\shortstack{Single\\ Image}}\\
\multirow{2}{*}{} & Source & Target & Train stage & Test stage & & \\
\hline
Unsupervised Domain Adaptation & $X^s, Y^s$  & $X^t$  & \checkmark	& - & \xmark & \xmark \\  
Test-Time Training & $X^s, Y^s$  & $X^t_\text{batch}$  & \checkmark	& \checkmark & \xmark & \xmark \\  
Fully Test-Time Adaptation & -  & $X^t_\text{batch}$  & -	& \checkmark & \xmark & \xmark \\  
Continual Test-Time Adaptation & -  & $X^t_\text{batch}$  & -	& \checkmark & \checkmark & \xmark \\  
\hline
Single Image Continual Test-Time Adaptation & -  & $x^t_i$  & -	& \checkmark  & \checkmark & \checkmark \\  
\bottomrule
\end{tabular}
}
\caption{ The settings of single image continual test-time adaptation and related adaptation areas. $X^s$ and $Y^s$ denote the source image and label sets, respectively, $X^t$ denotes the target image set, $X^t_\text{batch}=\{x^t_i, \cdots, x^t_{i+k}\}$ denotes the batch of $k$ target images, and $x^t_i$ denotes the single target image at time $i$.}
\label{tab:settings}
\end{table*}

Considering the nature of TTA, a TTA method must fulfill both the efficiency and accuracy requirements of real-world tasks. 
Meanwhile, real-world downstream vision tasks such as semantic segmentation require the use of high-resolution images for optimal performance~\cite{wang2020dual,zhao2018icnet}.  
These requirements limit the feasible batch size for edge computing environments where TTA methods are predominantly utilized, owing to the limited computational resources.

Due to the inherent limitations on batch size, existing works have yet to overcome the following two critical limitations, especially when using \textit{a single image} input, as shown in Table~\ref{tab:settings}.
Firstly, small mini-batch severely degrades the performances of most existing TTA methods based on batch normalization (BN)~\cite{wang2021tent,niu2022efficient,song2023ecotta,gong2022note,nado2021evaluating,wang2023dynamically,schneider2020improving}. 
These methods utilize BN to align the distributions between training and target data. However, inaccurate mini-batch statistics cause the substantial performance degradation when using a small batch size, which is more aggravated when dealing with a single image~\cite{niu2022towards_SAR,lim2022ttn}.
Secondly, entropy loss, which is employed by the majority of existing TTA methods~\cite{wang2021tent, song2023ecotta, lee2023towards_entropy,park2023label_entropy,gong2022note}, becomes significantly unstable when only a single image is utilized, due to the large and noisy gradients from the unreliable prediction~\cite{niu2022towards_SAR}. 
In addition, the entropy loss is prone to model collapse by predicting all images into the same class, which minimizes the entropy the most~\cite{niu2022towards_SAR}.
Although EATA~\cite{niu2022efficient} and  SAR~\cite{niu2022towards_SAR} proposed stabilized versions of entropy loss that filter unreliable samples, these methods still rely on the entropy and do not fully exploit the target samples due to the filtering.
Consistency loss using pseudo-labels \cite{wang2022continual, Brahma2023probabilistic} would overcome this problem, but it is inefficient in terms of computational complexity and memory consumption because it requires multiple inferences for tens of pseudo-labels.

In our paper, we present \textbf{BESTTA} (Beyond Entropy: Style Transfer guided single image cTTA), a single image cTTA method that achieves stable adaptation and computational efficiency. First of all, we formulate TTA as a style transfer problem from the target style to the source style. 
Motivated by the normalization-based style transfer methods~\cite{dumoulin2016learned_cin_style_transfer,huang2017arbitrary_adain}, we propose BeIN, a single normalization layer that enables stable style transfer of the input. 
BeIN estimates the proper mini-batch statistics describing the target domain by incorporating learnable parameters with the source and target input statistics. 
Moreover, we introduce the style and content losses to ensure effective and stable adaptation while preserving content. 
Our proposed losses outperform a single entropy loss in terms of stability and efficiency.
Second, we implement the style transfer in the simplest form for efficiency. We inject a single adaptive normalization layer, which has only two learnable parameters, into a model, and update a small number of parameters while keeping all model parameters frozen. 
Finally, we regularize the learnable parameters to prevent overfitting to the continually changing target domain, thereby lead to catastrophic forgetting and error accumulation.  


\par
Our contributions are summarized as follows:
\begin{itemize}
    \item We propose BESTTA, a novel style transfer guided single-image continual test-time adaptation method that enables stable and efficient adaptation. We formulate the test-time adaptation problem as a style transfer and propose novel style and content losses for stable single-image continual test-time adaptation.
    \item We propose a simple but powerful normalization method, namely BeIN. BeIN provides stable style transfer by learning to estimate the target statistics from input and source statistics.
    \item We demonstrate that the proposed BESTTA effectively adapts to continually changing target domains with a single image on semantic segmentation and image classification tasks. Remarkably, despite training only two parameters, BESTTA outperforms the existing state-of-the-art methods with the least memory consumption, as shown in Fig.~\ref{fig:teaser}.
    
    
    
    
\end{itemize}

\section{Related Work}


\subsection{Test-Time Adaptation}

In order to enable model adaptation in source-free, unlabeled, and online settings, various test-time domain adaptation methodologies~\cite{zhang2022memo,chen2022contrastivetta,niu2022towards_SAR,wang2021tent,niu2022towards_SAR,lim2022ttn,lee2023towards_entropy,park2023label_entropy} have been introduced, designing unsupervised losses for this purpose. TENT~\cite{wang2021tent} was the first to highlight the effectiveness of updating batch normalization based on entropy loss in a test-time adaptation task, inspiring subsequent studies to target batch normalization updates~\cite{niu2022efficient,song2023ecotta,gong2022note}. 
However, these methods require a large batch size for proper batch normalization statistics and show instability at small batch sizes~\cite{niu2022towards_SAR,lim2022ttn}. 
While some methods have aimed to facilitate TTA with small batch sizes~\cite{niu2022towards_SAR,lim2022ttn,lee2023towards_entropy,park2023label_entropy}, they often require prior knowledge of domain changes or auxiliary pretraining , making them impractical. 

To address this, continual TTA methods~\cite{song2023ecotta,wang2022continual,niu2022efficient,Brahma2023probabilistic,gong2022note} have emerged, aiming to prevent catastrophic forgetting and error accumulation caused by continued exposure to the inaccurate learning signal of unsupervised loss. To mitigate these issues, techniques such as stochastic parameter restoration~\cite{wang2022continual,Brahma2023probabilistic} and weight regularization losses~\cite{niu2022efficient} have been proposed. 
However, these methods also lack of considerations for a small batch sizes during adaptation.


\subsection{Style Transfer}
Neural style transfer methods have been proposed to transfer the style of the image while preserving the content~\cite{gatys2016image_style_transfer,li2017universal_style_transfer,li2017demystifying_style_transfer,ulyanov2016texture_style_transfer,gal2022stylegannada_style_transfer,zhang2018separating_style_transfer,gatys2017controlling_style_transfer}.
Among them, some studies proposed normalization techniques to effectively transfer the image~\cite{dumoulin2016learned_cin_style_transfer,huang2017arbitrary_adain}. Using a similar method, Fahes \textit{et al.}~\cite{fahes2023poda} transferred the input styles in a dataset to the target style and trained subsequent neural networks for downstream tasks. Motivated by these methods, we propose to transfer the target feature to the source domain, updating the normalization layer in the cTTA setting.


\begin{figure*}[ht!]
    \centering
    \includegraphics[width=\linewidth]{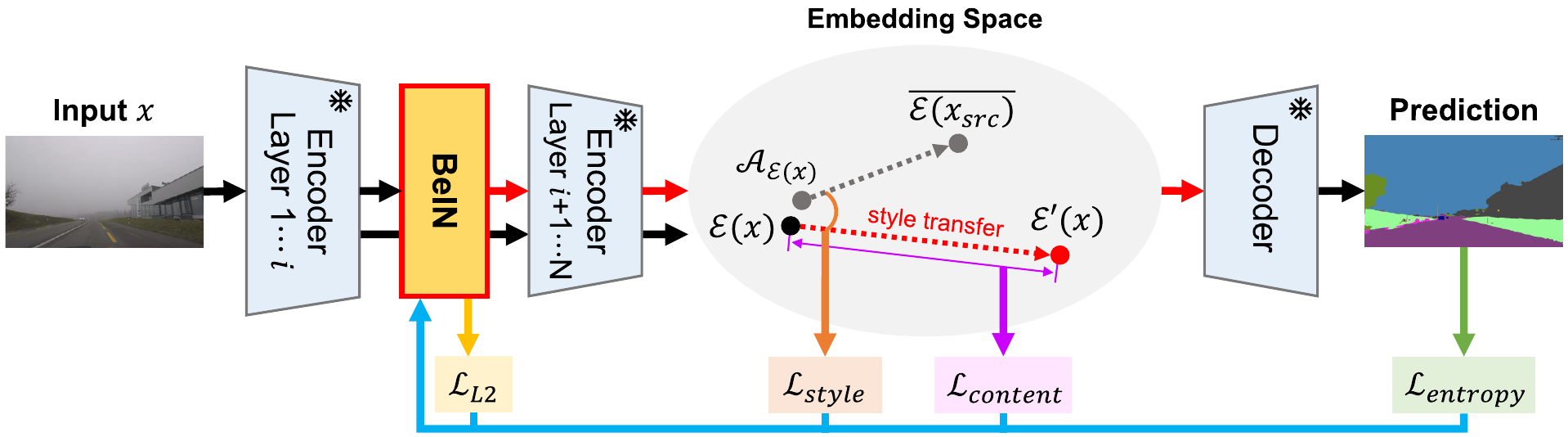}
    \caption{Overview of BESTTA. We inject a single normalization layer, BeIN, to the pretrained model, which facilitates the stable single image continual test-time adaptation via style transfer. $\mathcal{E}(x)$ denotes the input embedding, $\mathcal{E}'(x)$ denotes the adapted input embedding, $\overline{\mathcal{E}(x_{src})}$ denotes the mean source embedding, and $\mathcal{A}_{\mathcal{E}(x)}$ denotes the exponential moving average of the target embeddings.
    To transfer the style of the input image to the source style, we adopt the directional style loss $\mathcal{L}_{style}$. To avoid the distortion of the content in the input image, we take the content loss $\mathcal{L}_{content}$. And to further improve the performance of the model, we use the entropy loss, i.e., $\mathcal{L}_{entropy}$. Finally, to avoid catastrophic forgetting and error accumulation, we use the L2 loss $\mathcal{L}_{L2}$. During adaptation, only two normalization parameters in BeIN are trained.}
    \label{fig:arch}
\end{figure*}

\section{Method}
Fig.~\ref{fig:arch} illustrates the overview of the proposed method, BESTTA. 
Motivated by style transfer methods~\cite{dumoulin2016learned_cin_style_transfer,huang2017arbitrary_adain} that use a normalization layer, we formulate the TTA problem as a style transfer problem in Sec~\ref{sec:probdef}. We propose a single normalization layer, BeIN, which stabilizes the target input statistics in single image cTTA in Sec~\ref{sec:bein}, with the proposed losses in Sec~\ref{sec:loss}.


\subsection{Problem Definition}
\label{sec:probdef}
\topic{Stable adaptation on a single image.}
Batch normalization (BN) \cite{ioffe2015batchnorm} utilizes mini-batch statistics, mean and standard deviation of the source data to normalize data on the same distribution for each channel:
\begin{equation}
    \text{BN}(X) = \alpha \cdot \frac{X - \mu_{s}}{\sigma_{s}} + \beta
    \label{eq:bn}
\end{equation}
where $X \in \mathbb{R}^{B\times C\times H\times W}$ denotes the mini-batch of input features with $C$ channels, $\alpha, \beta \in \mathbb{R}^C$ denote learnable affine parameters optimized during pretraining, and the mean $\mu_s \in \mathbb{R}^C$ and the standard deviation $\sigma_s \in \mathbb{R}^C$ are obtained by exponentially averaging the features during pretraining.



TENT~\cite{wang2021tent} introduces the concept of updating affine parameters of BN layers. TENT replaces the source statistics $(\mu_s, \sigma_s)$ to a target statistics $(\mu_X, \sigma_X)$ of a target input feature $X$ to address the distribution shift, and trains the learnable affine parameters $\alpha$ and $\beta$ during test-time:
\begin{equation}
    \text{BN}_{\text{TENT}}(X) = {\alpha} \cdot \frac{X - \mu_X}{\sigma_X} + {\beta}
    \label{eq:tent}
\end{equation}
The parameters are updated with the entropy loss $H(\hat{y}) = - \Sigma_c p(\hat{y}) \log{p(\hat{y})}$, where $\hat{y}$ is the prediction of the target input.
This approach and related studies~\cite{niu2022efficient,song2023ecotta} have demonstrated significant promise in TTA with the large batch size. 

However, the normalization-based methods suffer significant performance drop when using small mini-batches, because they depend on the assumption that the true mean $\mu_t$ and variance $\sigma_t^2$ of the target distribution can be estimated by the sample statistics. By the central limit theorem, the sample mean $\mu_X$ follows the Gaussian distribution $N(\mu_t, \frac{\sigma_t^2}{n})$ for sufficient large samples of size $n$. The sample variance follows the Chi-square distribution such that $(n-1)\frac{\sigma_X^2}{\sigma_t^2} \sim \chi^2_{n-1}$, and the variance of this distribution $Var(\sigma_X^2) = \frac{2\sigma_t^4}{n-1}$, where the data are normally distributed. Since the variances of the distributions of both sample statistics significantly increase when $n$ is small, estimating the true mean and variance based on the sample mean and variance becomes difficult, especially for a single image. Therefore, estimating the true mean and variance accurately is required to ensure reliable adaptation with a single image.
Also, the entropy loss utilized in these methods is unstable because the instability of BN persists due to its linearity. Furthermore, the entropy loss often results in model collapse, causing biased prediction~\cite{niu2022towards_SAR}. 

Therefore, when dealing with a single image, a solution is required that (1) stabilizes the input statistics in the normalization layers and (2) uses stable losses more than minimizing entropy.

\topic{Test-time adaptation as a style transfer.}
In the style transfer domain, there have been several methods that utilize normalization layers for style transfer~\cite{dumoulin2016learned_cin_style_transfer,huang2017arbitrary_adain}.
Dumoulin \textit{et al.}~\cite{dumoulin2016learned_cin_style_transfer} proposed a conditional instance normalization (CIN) that trains the learnable parameters $\alpha^s$ and $\beta^s$ to transfer the style of the encoded input $x$ to the style $s$:
\begin{equation}
    \text{CIN}(x;s) = \alpha^s \cdot \frac{x - \mu_x}{\sigma_x} + \beta^s
    \label{eq:objective}
\end{equation}
Similarly, AdaIN~\cite{huang2017arbitrary_adain} uses the target style directly in its instance normalization to transfer the style of encoded input $x$ to the style of encoded target style input $y$:

\begin{equation}
    \text{AdaIN}(x, y) = \sigma_y \cdot \frac{x - \mu_x}{\sigma_x} + \mu_y
    \label{eq:objective}
\end{equation}
where $\sigma_y$ and $\mu_y$ denote the standard deviation and mean of the encoded target style input $y$.
Despite their simplicity, they have shown promising results in style transfer.
However, it is important to note that all inputs utilized in the aforementioned methods are in-distribution, which means that the models are trained on both source and target data, therefore the statistics of the encoded input image $(\mu_x, \sigma_x)$ are reliable. 
In contrast, cTTA setting involves inputs from an arbitrary target domain that are out-of-distribution, resulting in unreliable and unstable statistics.
Motivated by this, we formulate the TTA as a style transfer problem that transfers the target style to the source style:
\begin{equation}
    \text{TTA}(x)=\sigma_s \cdot \frac{x - \mu_t}{\sigma_t} + \mu_s
    \label{eq:tta_as_style}
\end{equation}
where the source and the target domain follow the distributions $N(\mu_s, \sigma_s^2)$ $N(\mu_t, \sigma_t^2)$, respectively.
\subsection{BeIN: BESTTA Instance Normalization}
\label{sec:bein}

We propose a BESTTA Instance Normalization (BeIN) layer that  transfers the style of a target input to the source style, enabling seamless operation of the latter parts of a model.
For stability, we estimate the target statistics $(\mu_t, \sigma_t)$ using an anchor point and learnable parameters $\gamma_\sigma$ and $\gamma_\mu$. We use the source style $(\overline{\mu_s}, \overline{\sigma_s})$ as the anchor point because it is fixed and therefore stable. The source style contains the mean $\overline{\mu_s}$ and the standard deviation $\overline{\sigma_s}$ of the source features, which are small and can be easily obtained from the training phase, before the deployment of our method. BeIN is formulated as:
\begin{equation}
    \text{BeIN}(x)=\overline{\sigma_s} \cdot \frac{x - \hat\mu_t}{\hat\sigma_t} + \overline{\mu_s}
    \label{eq:objective}
\end{equation}
where $\hat\mu_t$ and $\hat\sigma_t$ denote the estimated target mean and standard deviation, respectively. We estimate the true target statistics by combining the target input statistics and the source style. We estimate $\hat\sigma_t$ as a weighted harmonic mean of the source standard deviation and the target input standard deviation with a learnable parameter $\gamma_\sigma$:

\begin{equation}
    \hat\sigma_t = \frac{\overline{\sigma_s}\cdot\sigma_x}{\rho\overline{\sigma_s} + (1-\rho)\sigma_x + \gamma_{\sigma}}
    \label{Eq:sigma}
\end{equation}
where $\rho$ is the hyperparameter that adjusts the ratio of using the source statistics.
For the mean, we estimate it by the weighted sum of the source mean and the target input mean with a learnable parameter $\gamma_\mu$:

\begin{equation}
    \hat\mu_t 
    = \rho \frac{\hat\sigma_t}{\sigma_x}\cdot\mu_x + (1-\rho)\frac{\hat\sigma_t}{\overline{\sigma_s}}\cdot\overline{\mu_s} + \gamma_\mu 
    \label{Eq:mu}
\end{equation}
The means are scaled with the standard variations to be aligned.
We insert BeIN between the layers of the encoder as depicted in Fig.~\ref{fig:arch}. By training only a two parameters in the embedded layer, the latter part of the model seamlessly operates with the preserved prebuilt knowledge, leading to efficient learning and modularization of the whole model.

\begin{figure}[t]
    \centering
    \includegraphics[width=\linewidth]{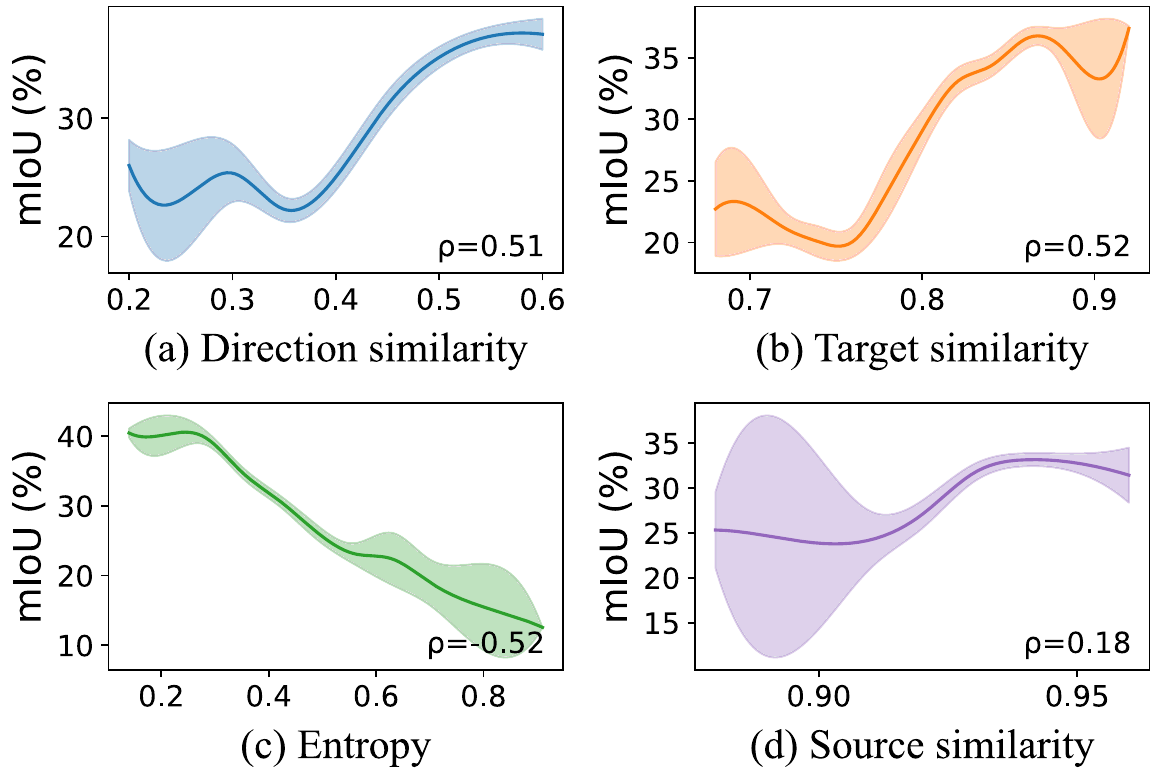}
    \caption{Correlations between performance and style transfer related metrics.
    We find that direction similarity and target similarity have high correlation ($\rho > 0.5$) with the performance, whereas source similarity~\cite{patashnik2021styleclip, fahes2023poda} is uncorrelated ($\rho = 0.18$). We use the method of Schneider \textit{et al.}~\cite{schneider2020improving} to measure the similarity between the adapted and unadapted embeddings. All results are evaluated on the ACDC dataset~\cite{sakaridis2021acdc} using DeepLabV3Plus-ResNet50~\cite{chen2018encoder} pretrained on the Cityscapes dataset~\cite{cordts2016cityscapes}.
    }
    \label{fig:loss_correlation}
\end{figure}

\subsection{Style-guided losses}
\label{sec:loss}
Conventional style losses~\cite{li2017demystifying_style_transfer,huang2017arbitrary_adain,dumoulin2016learned_cin_style_transfer,gatys2016image_style_transfer} can be used to guide the learnable parameters in BeIN to transfer the style of the input features to the source style. These style losses align the distribution of the transferred image to the target style. However, they require heavy computations such as additional encoding or decoding processes or computing the gram matrix. Therefore, considering the efficiency requirement of  TTA methods, adopting the conventional style losses is impractical.
To address this problem, we propose novel directional style loss and content loss that are computed in the embedding space of encoder during the inference, satisfying efficiency without heavy operations.

\topic{Directional style loss.}
Conventional style transfer methods utilize the similarity between an encoded input and an encoded style input. The straightforward solution to our method is to measure the source similarity $\cos(\overline{\mathcal{E}(x_{src})}, \mathcal{E}'(x))$~\cite{fahes2023poda, patashnik2021styleclip}, that is, the cosine similarity between the adapted target embedding $\mathcal{E}'(x)$ and the mean source embedding $\overline{\mathcal{E}(x_{src})}$. However, as shown in Fig.~\ref{fig:loss_correlation}d, we empirically find that the source similarity is not correlated with the performance, thus not beneficial to improve the adaptation. Therefore, motivated by StyleGAN-NADA~\cite{gal2022stylegannada_style_transfer}, we devise the directional style loss. We find that the similarity between the adaptation direction $(\overline{\mathcal{E}(x_{src})} - \mathcal{A}_{\mathcal{E}(x)})$ and the direction from the target to the source $(\mathcal{E}'(x) - \mathcal{E}(x))$ (see Fig.~\ref{fig:arch}) has high correlation with the adaptation performance as illustrated in Fig.~\ref{fig:loss_correlation}a. Therefore, we formulate our directional style loss as follows:
\begin{equation}
    \mathcal{L}_{style} = 1 - \cos((\overline{\mathcal{E}(x_{src})} - \mathcal{A}_{\mathcal{E}(x)}) , (\mathcal{E}'(x) - \mathcal{E}(x)))
    \label{Eq:styleloss}
\end{equation}
where $\mathcal{E}(x)$ is the unadapted target embedding, $\mathcal{A}_\mathcal{E}(x)$ is the exponential moving average of target embeddings.


\topic{Content loss.}
Style transfer without consideration about the content of the transferred feature leads to distortion of the contents~\cite{li2017demystifying_style_transfer,gatys2016image_style_transfer,huang2017arbitrary_adain,zhang2018separating_style_transfer}. Similar to these findings, we find that the target similarity, that is, the cosine similarity between the transferred feature $\mathcal{E}'(x)$ and the input feature $\mathcal{E}(x)$ has high correlation with the performance, as shown in Fig.~\ref{fig:loss_correlation}b. Therefore, we devise the content loss as follows:
\begin{equation}
    \mathcal{L}_{content} = 1 - \cos({\mathcal{E}(x)} ,  \mathcal{E}'(x))
    \label{Eq:contentloss}
\end{equation}

\topic{L2 regularization.}
In the continual TTA setting, where long-term adaptation is necessary, it is crucial to prevent catastrophic forgetting and error accumulation to ensure optimal performance.  
Therefore, we employ an L2 norm to regularize the learnable parameters to prevent overfitting to the current target domain. We formulate the L2 regularization loss as follows:
\begin{equation}
    \mathcal{L}_{L2} = \|\gamma_{\mu}\|_2 + \|\gamma_{\sigma}\|_2
    \label{Eq:contentloss}
\end{equation}

\topic{Entropy loss.}
We incorporate the entropy loss introduced by Wang \textit{et al.}~\cite{wang2021tent}, as BeIN stabilize the normalization. We verify that entropy has strong correlation with the performance on semantic segmentation task as shown in Fig. \ref{fig:loss_correlation}c. The entropy loss is as follows:
\begin{equation}
\mathcal{L}_{entropy} = - \Sigma p(\hat{y}) \log{p(\hat{y})}
\end{equation}
where $p$ denotes probability, $\hat{y}$ denotes prediction.


\topic{Total loss.} We train the learnable parameters in our proposed BeIN layer with a combination of the proposed losses. The total loss $\mathcal{L}$ is as follows:
\begin{equation}
    \mathcal{L} = \lambda_1 \cdot \mathcal{L}_{style} + \lambda_2 \cdot \mathcal{L}_{content} + \lambda_3 \cdot \mathcal{L}_{entropy} + \lambda_4 \cdot \mathcal{L}_{L2}
    \label{Eq:totalloss}
\end{equation}
where $\lambda_1$, $\lambda_2$, $\lambda_3$, and $\lambda_4$ are the weights for each loss.



\begin{table*}[t!]
\centering
\def\arraystretch{1.1}
\setlength{\tabcolsep}{1pt}
{\small
\begin{tabularx}{\textwidth}{c|
>{\centering\arraybackslash}m{1cm}
>{\centering\arraybackslash}m{1cm}
|*{4}{YYYY|}c}
\hline
\toprule 
 \multicolumn{3}{c|}{Time} & \multicolumn{16}{l|}{$\; t \; \xrightarrow{\hspace{1.31\columnwidth}}$} & \multirow{3}{*}{Mean}  \\
\cline{1-19}
 \multirow{2}{*}{Method} & 
 \multirow{2}{*}{\makecell{Memory\\(GB)}} & \multirow{2}{*}{\makecell{Time\\(ms)}} 
 & \multicolumn{4}{c|}{Round 1} &  \multicolumn{4}{c|}{Round 4} &  \multicolumn{4}{c|}{Round 7} &  \multicolumn{4}{c|}{Round 10} &  \\
 &  && Fog & Night & Rain & Snow & Fog & Night & Rain & Snow & Fog & Night & Rain & Snow & Fog & Night & Rain & Snow &   \\
\hline
Source & 1.76 & 135.9 & 44.3 & 22.0 & 41.5 & 38.9 & 44.3 & 22.0 & 41.5 & 38.9 & 44.3 & 22.0 & 41.5 & 38.9 & 44.3 & 22.0 & 41.5 & 38.9 & 36.6 \\
\hline
BN Stats Adapt~\cite{nado2021evaluating} & 2.01 & 192.7 & 36.8 & \underline{23.5} & 38.2 & 36.3 & 36.8 & 23.5 & 38.2 & 36.3 & 36.8 & 23.5 & 38.2 & 36.3 & 36.8 & 23.5 & 38.2 & 36.3 & 33.7 \\
TENT$^*$~\cite{wang2021tent} & 9.58 & 343.8 & 38.2 & 22.9 & 41.1 & 37.8 & 38.2 & 22.9 & 41.1 & 37.8 & 38.2 & 22.9 & 41.1 & 37.8 & 38.2 & 22.9 & 41.1 & 37.8 & 35.0 \\
TENT-continual~\cite{wang2021tent} & 9.58 & 343.8 & 37.7 & \underline{23.5} & 39.9 & 37.5 & 31.4 & 17.3 & 30.7 & 26.8 & 19.8 & 11.2 & 18.8 & 17.3 & 13.4 & 8.7 & 12.4 & 11.8 & 22.3 \\
EATA~\cite{niu2022efficient} & 9.90 & 190.6 & 36.8 & 23.4 & 38.3 & 36.5 & 37.6 & \underline{23.5} & 39.0 & 37.1 & 38.1 & \underline{23.6} & 39.5 & 37.6 & 38.5 & \underline{23.6} & 39.9 & 38.0 & 34.4 \\
CoTTA~\cite{wang2022continual} & 18.20 & 4337 & \underline{46.3} & 22.0 & \underline{44.2} & 40.4 & \underline{48.1} & 21.0 & \underline{45.3} & 40.2 & \underline{48.1} & 20.4 & \underline{45.3} & 39.7 & \underline{48.0} & 20.0 & \underline{45.1} & 39.5 & 38.4 \\
EcoTTA~\cite{song2023ecotta} & 10.90 & 759.3 & 33.4 & 20.8 & 35.6 & 32.9 & 34.2 & 21.2 & 36.1 & 33.2 & 34.7 & 21.3 & 36.3 & 33.2 & 34.9 & 21.4 & 36.5 & 33.2 & 31.2 \\
PETAL~\cite{Brahma2023probabilistic} & 18.51 & 5120 & 44.7 & 22.1 & 42.9 & 40.1 & 47.3 & 22.4 & 44.4 & \underline{40.8} & 47.1 & 22.1 & 44.7 & \underline{40.6} & 46.9 & 22.0 & 44.6 & \underline{40.5} & \underline{38.5} \\
\rowcolor{lightgray}
BESTTA (Ours) & {8.77} & 291.8 & \textbf{47.8} & \textbf{24.3} & \textbf{47.2} & \textbf{43.8} & \textbf{54.5} & \textbf{26.1} & \textbf{48.4} & \textbf{45.2} & \textbf{54.5} & \textbf{26.1} & \textbf{48.4} & \textbf{45.3} & \textbf{54.5} & \textbf{26.1} & \textbf{48.4} & \textbf{45.3} & \textbf{43.2} \\[-0.3ex]
\bottomrule
\end{tabularx}
}
\caption{Semantic segmentation results (mIoU in \%) on Cityscapes-to-ACDC single image continual test-time adaptation task. We compare ours with other state-of-the-art TTA methods in terms of peak GPU memory usage (GB) and time consumption (ms) for each iteration, and mean intersection over union (mIoU). All results are evaluated using the DeepLabV3Plus-ResNet50. $^*$ denotes the requirement about when the domain shift occurs. The \textbf{best} and \underline{second best} results are highlighted. }
\label{tab:SS_ACDC}
\end{table*}

\begin{table}[t!]
\centering
\small
\def\arraystretch{1}
\setlength{\tabcolsep}{3pt}
\begin{tabularx}{\linewidth}{c|*{4}{Y}|c}
\hline
\toprule
Time & \multicolumn{4}{l|}{$\; t \xrightarrow{\hspace{0.44\columnwidth}}$} & \multirow{2}{*}{Mean}  \\
\cline{1-5}
Method & Bright. & Fog & Frost & Snow  \\
\hline
Source & 67.5 & 59.3 & 24.8 & 13.9 & 41.4 \\
TENT-continual~\cite{wang2021tent} & 61.9 & 45.2 & 23.0 & 15.3 & 36.3 \\
PETAL~\cite{Brahma2023probabilistic} & 64.7 & 35.8 & 5.7 & 0.4 & 26.7 \\
\rowcolor{lightgray}
BESTTA (Ours) & \textbf{68.2} & \textbf{60.5} & \textbf{31.6} & \textbf{20.0} & \textbf{45.0} \\[-0.3ex]
\bottomrule
\end{tabularx}
\caption{Semantic segmentation results (mIoU in \%) on the Cityscapes-to-Cityscapes-C gradual test-time adaptation task. All experiments are evaluated using DeepLabV3Plus-ResNet50.}
\label{tab:main-cityscapes-c-gradual}
\end{table}

\begin{figure*}[t]
    \centering
    \includegraphics[width=0.85\linewidth]{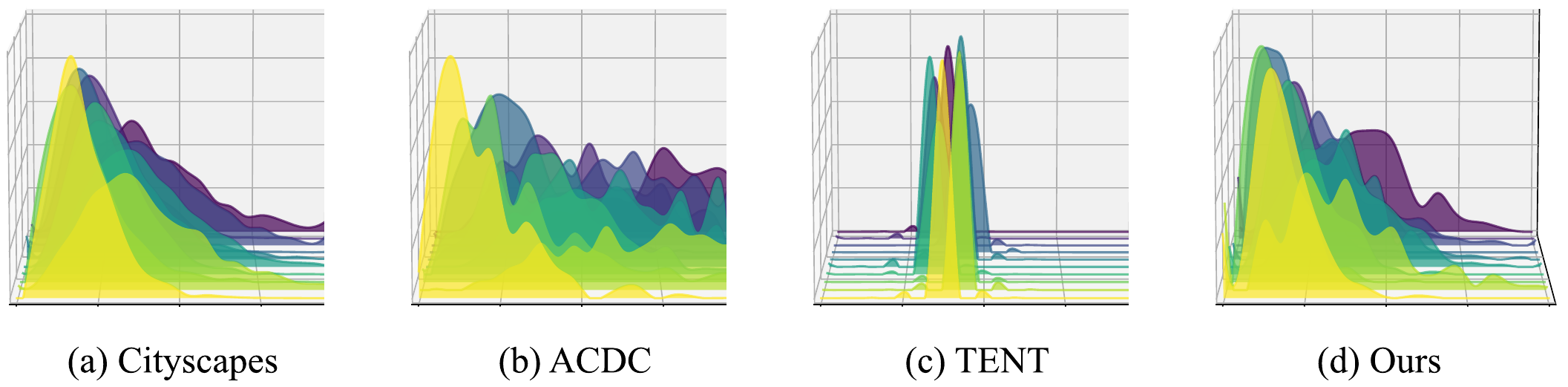}
    \caption{Embedding feature distributions of (a) Cityscapes dataset on Source model, (b) ACDC dataset on Source model, (c) ACDC dataset on TENT, (d) ACDC dataset on our method. Ours aligns the distribution with the source dataset, whereas TENT collapses into a single point. We use DeepLabV3Plus-ResNet50 pretrained on Cityscapes in this experiment.}
    \label{fig:distribution}
\end{figure*}

\begin{figure*}[t]
    \centering
    \includegraphics[width=0.95\linewidth]{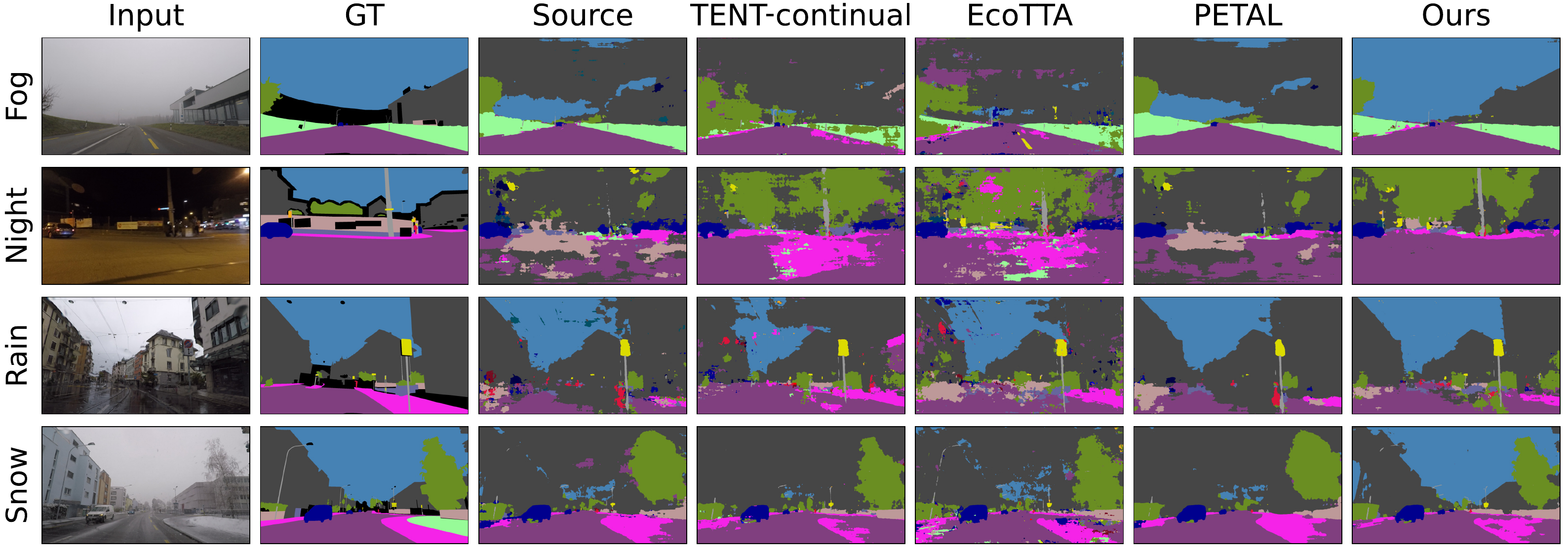}
    \caption{Qualitative comparison of semantic segmentation on Cityscapes-to-ACDC task. All experiments are evaluated using DeepLabV3Plus-ResNet50. The results of other methods are presented in the supplementary material.}
    \label{fig:Qual_SS}
\end{figure*}

\begin{table*}[t!]
\centering
\small
\def\arraystretch{1}
\setlength{\tabcolsep}{3pt}
\def\tabmid{\centering\arraybackslash}
\def\losslsp{1.5cm}
\def\lossrsp{1.4cm}
\begin{tabular}{>{\tabmid}m{\losslsp} >{\tabmid}m{\losslsp} >{\tabmid}m{\losslsp} >{\tabmid}m{\losslsp}
            |   >{\tabmid}m{\lossrsp} >{\tabmid}m{\lossrsp} >{\tabmid}m{\lossrsp} >{\tabmid}m{\lossrsp}
            |   >{\tabmid}m{\lossrsp} }
\hline
\toprule
$\mathcal{L}_{entropy}$ & $\mathcal{L}_{style}$ & $\mathcal{L}_{content}$ & $\mathcal{L}_{L2}$ & {Round 1} & {Round 4} & {Round 7} & {Round 10} & {Mean}\\
\hline 
\checkmark & \xmark & \xmark & \xmark  & 21.2 & 5.3 & 4.5 & 4.2 & 6.7 \\
\checkmark & \checkmark& \xmark & \xmark  & 37.1 & 32.3 & 30.2 & 28.9 & 31.8  \\
\checkmark & \checkmark & \checkmark & \xmark  & 40.1 & 39.2 & 37.9 & 37.3 & 38.6 \\
\rowcolor{lightgray}
\checkmark & \checkmark & \checkmark & \checkmark  & \textbf{40.8} & \textbf{43.5} & \textbf{43.6} & \textbf{43.6} & \textbf{43.2}  \\[-0.3ex]
\bottomrule
\end{tabular}
\caption{Ablation study on our losses. Performances (mIoU in \%) are evaluated on Cityscapes-to-ACDC task for each different combination of losses, using DeepLabV3Plus-ResNet50~\cite{chen2018encoder}.}
\label{tab:ablation_loss}

\end{table*}

\begin{table}[ht!]
\centering
\small
\def\arraystretch{1}
\setlength{\tabcolsep}{5pt}
\begin{tabularx}{0.75\linewidth}{YYYY|c}
\hline
\toprule
\textit{Layer1} & \textit{Layer2} & \textit{Layer3} & \textit{Layer4} & \textit{mIoU}\\
\hline
\checkmark & \xmark & \xmark & \xmark & 41.7 \\
\xmark  & \checkmark  & \xmark & \xmark & 40.9 \\
\rowcolor{lightgray}
\xmark  & \xmark & \checkmark & \xmark & \textbf{43.2}\\
\xmark  & \xmark  & \xmark & \checkmark & 36.0 \\
\hline
\checkmark & \xmark  & \checkmark & \xmark & 43.0 \\ 
\xmark & \checkmark  & \checkmark & \xmark & 41.7 \\
\xmark & \xmark  & \checkmark & \checkmark & 41.4 \\ 
\bottomrule
\end{tabularx}
\caption{Ablation study on the position of BeIN. Performance (mIoU in \%) are evaluated on Cityscapes-to-ACDC task, using DeepLabV3Plus-ResNet50.}
\label{tab:abalation_layer}
\end{table}

\newcommand{\rota}[1]{\rotatebox[origin=c]{60}{\small{#1}}}

\begin{table*}[t!]
\centering
\def\arraystretch{1.1   }
\setlength{\tabcolsep}{3pt}
\small {
\begin{tabularx}{\textwidth}{c|*{15}{Y}|Y}
\hline
\toprule
Time & \multicolumn{15}{l|}{$\; t \xrightarrow{\hspace{1.55\columnwidth}}$} & \multirow{2}{*}{Mean}  \\
\cline{1-16}
Method & \rota{Gaussian} &  \rota{shot} &  \rota{impulse} &  \rota{defocus} &  \rota{glass} &  \rota{motion} &  \rota{zoom} &  \rota{snow} &  \rota{frost} &  \rota{fog} &  \rota{brightness} &  \rota{contrast} &  \rota{elastic\_trans} &  \rota{pixelate} &  \rota{jpeg}  & \\
\hline
Source & 88.9 & 89.2 & 89.6 & 77.5 & 83.2 & \underline{75.6} & \underline{77.2} & 75.2 & 70.0 & \underline{68.4} & 66.9 & \underline{49.9} & 81.0 & 81.2 & 83.2 & 77.1 \\
BN Stats Adapt~\cite{nado2021evaluating}  & 89.5 & 89.4 & 89.7 & 90.1 & 89.5 & 89.5 & 89.5 & 89.6 & 89.6 & 89.2 & 89.8 & 90.1 & 89.5 & 89.4 & 89.6 & 89.6 \\
TENT-continual~\cite{wang2021tent} & 90.0 & 90.0 & 90.0 & 90.0 & 90.0 & 90.0 & 90.0 & 90.0 & 90.0 & 90.0 & 90.0 & 90.0 & 90.0 & 90.0 & 90.0 & 90.0 \\
CoTTA~\cite{wang2022continual} & \underline{54.0} & 64.0 & 78.0 & 86.8 & 88.0 & 89.9 & 89.7 & 87.8 & 87.6 & 89.7 & 88.7 & 90.0 & 89.4 & \underline{53.9} & 86.7 & 81.6 \\

PETAL~\cite{Brahma2023probabilistic} & \textbf{51.1} & \underline{52.0} & \underline{77.7} & \underline{73.4} & \underline{62.8} & 79.7 & 84.1 & \underline{48.3} & \underline{67.4} & 78.9 & \underline{19.0} & 81.5 & \underline{56.5} & 57.4 & \underline{69.8} & \underline{64.0} \\

\rowcolor{lightgray}
BESTTA (Ours) & {55.2} & \textbf{50.5} & \textbf{65.6} & \textbf{38.1} & \textbf{46.1} & \textbf{29.3} & \textbf{32.6} & \textbf{23.4} & \textbf{33.1} & \textbf{21.8} & \textbf{9.1} & \textbf{38.5} & \textbf{24.5} & \textbf{44.3} & \textbf{28.9} & \textbf{36.1} \\[-0.3ex]
\bottomrule
\end{tabularx}
}
\caption{Image classification results on CIFAR-10-C. All results are evaluated using WideResNet-28 pre-trained on CIFAR-10. We use the error rate (\%) as the metric. The \textbf{lowest} and \underline{second lowest} error rates are highlighted.}
\label{tab:CLF_CIFAR10}
\end{table*}

\section{Experiments}
\label{sec:exp}
We conduct two experiments on our proposed method 
in terms of 
semantic segmentation and image classification. We use the following methods as baselines:
\begin{itemize}
    \item Fully TTA: BN Stats Adapt~\cite{nado2021evaluating} and TENT~\cite{wang2021tent}
    \item Continual TTA: TENT-continual~\cite{wang2021tent}, EATA~\cite{niu2022efficient}, CoTTA~\cite{wang2022continual}, EcoTTA~\cite{song2023ecotta}, and PETAL~\cite{Brahma2023probabilistic}
\end{itemize}

\subsection{Experiments on Semantic Segmentation}
We evaluate our proposed method and other baselines in two different settings on semantic segmentation: Cityscapes-to-ACDC continual TTA and Cityscapes-to-Cityscapes-C gradual TTA. All semantic segmentation results are evaluated in mean intersection over union (mIoU).

\label{sec:ExpSS}
\topic{Experimental setup.} 
Following the setting of CoTTA~\cite{wang2022continual}, we conduct experiments in the continually changing target environment. We adopt the Cityscapes dataset~\cite{cordts2016cityscapes} as the source dataset and the Adverse Conditions (ACDC) dataset~\cite{sakaridis2021acdc} as the target dataset. The ACDC dataset includes four different adverse weather conditions (i.e., fog, night, rain, and snow) captured in the real-world. We conduct all experiments without the domain label except for TENT~\cite{wang2021tent}. We repeat the same sequence of four weather conditions 10 times (i.e., fog $\rightarrow$ night $\rightarrow$ rain $\rightarrow$ snow $\rightarrow$ fog $\rightarrow$ $\cdots$, 40 conditions in total). 

We also conduct experiments in the gradually changing target environment to simulate more realistic environments.
We adopt the Cityscapes dataset as the source dataset and the Cityscapes-C dataset~\cite{michaelis2019cityscapes_c,hendrycks2018benchmarking} as the target dataset. The Cityscapes-C dataset is the corrupted version of the Cityscapes dataset that includes 5 severity levels and 15 types of corruptions. Following EcoTTA~\cite{song2023ecotta}, we utilize the four most realistic types of corruptions (i.e., brightness, fog, frost, snow) in our experiment. The model faces varying levels of corruption severity for a specific weather type, progressing from 1 to 5 and then from 5 to 1. Once the severity reaches the lowest level, the corruption type is changed (e.g., brightness 1 $\rightarrow$ 2 $\rightarrow \cdots \rightarrow$ 5 $\rightarrow$ 4 $ \cdots\rightarrow$ 1 $\rightarrow$ fog 1$\rightarrow$ 2 $\rightarrow \cdots$, 36 conditions in total).

\topic{Implementation Details.} 
We utilize ResNet50-DeepLabV3+~\cite{chen2018deeplabv3+} pretrained on the Cityscapes dataset. 
We set the batch size to 1 with an image of size $1920\times1080$ for the continually changing setting and a $2048\times1024$ size for the gradually changing setting, respectively.
We collect the source style $(\overline{\mu_s}, \overline{\sigma_s})$ and the mean source embedding $\overline{\mathcal{E}(x_{src})}$ by inferencing the source data before deployment.
We insert the proposed BeIN layers between the third and fourth layers in the backbone. The $\lambda_1$, $\lambda_2$, $\lambda_3$, and $\lambda_4$ are set to 0.3, 1.0, 0.3, and 0.04, respectively. For optimization, we employ the SGD optimizer with a learning rate of 0.001 for the adaptation phase. Note that the model was pretrained with a learning rate of 0.1. The experiments are conducted using an NVIDIA RTX3090 GPU.


\topic{Quantitative Results.} As shown in Table~\ref{tab:SS_ACDC}, we compare our approach with the baselines on the cTTA setting. Our proposed BESTTA achieves the highest mIoU across all weather types and rounds. 
In particular, our method significantly outperforms the second-best baseline, PETAL~\cite{Brahma2023probabilistic}, with a mean mIoU gap of 4.7\%, despite using 5.7\% of the processing time per image compared to the second-best model. 
Ours also prevent catastrophic forgetting and error accumulation in the long-term adaptation, as ours shows consistently high performance.
In contranst, the BN-based methods~\cite{wang2021tent,niu2022efficient,song2023ecotta} show severe performance degradation.
Notably, even we reinitialize the TENT reinitialize whenever each domain change occurs, the performance is lower than the source model.
This indicates that the unstable mini-batch statistics and entropy loss hinder the adaptation in single image cTTA.


The experimental results for the gradually changing setting are presented in Table~\ref{tab:main-cityscapes-c-gradual}. 
Similar to the cTTA setting, our method significantly outperforms the baselines. 
All the baselines perform worse than the source model, indicating that they fail to deal with the single image cTTA setting.

\topic{Qualitative Results.} 
Fig.~\ref{fig:distribution} provides the distributions of the adapted features of ours and the baselines. Comparing Fig.~\ref{fig:distribution}a and Fig.~\ref{fig:distribution}b, the feature distribution of the target domain that is obtained by the source model is diversified. The distribution of TENT-adapted feature is collapsed into a single point, due to the entropy loss. In contrast, our adapted feature is aligned with the source features from the source model. This indicates that our BESTTA effectively and stably adapts the model to the target domain.

Fig.~\ref{fig:Qual_SS} shows predictions of ours and the baselines on the ACDC dataset. 
Compared to the other state-of-the-art methods, our results are clearer and more accurate. In particular, at night, the baselines show globally noisy predictions while ours show clear results; in fog and snow, our method perceives the sky as sky while others predict it as buildings. These observations demonstrate the effectiveness of our method in a real-world environment.
Further results are presented in the supplementary material. 


\topic{Effectiveness of losses.}
We perform an ablation study on our losses, and the results are provided in Table~\ref{tab:ablation_loss}. Updating the parameters in BeIN only with entropy losses yields the lowest performance. Conversely, including style loss significantly improves performance, demonstrating the effectiveness of style transfer in our method. The addition of content loss further improves performance by preserving the content of the input image, allowing only feature styles to be transferred. However, it still exhibits error accumulation, as evidenced by a gradual performance degradation. L2 loss effectively prevents this phenomenon by maintaining stable performance across rounds.

\topic{Selection of layer to insert BeIN layer.}
Table~\ref{tab:abalation_layer} shows the effectiveness of the selection of layers, where the BeIN layer is inserted to transfer the features. Adaptation of features from layer third achieves the highest performance. However, the addition of other layers leads to a performance degradation compared to using only the third layer.  

\begin{figure}[t]
    \centering
    \includegraphics[width=\linewidth]{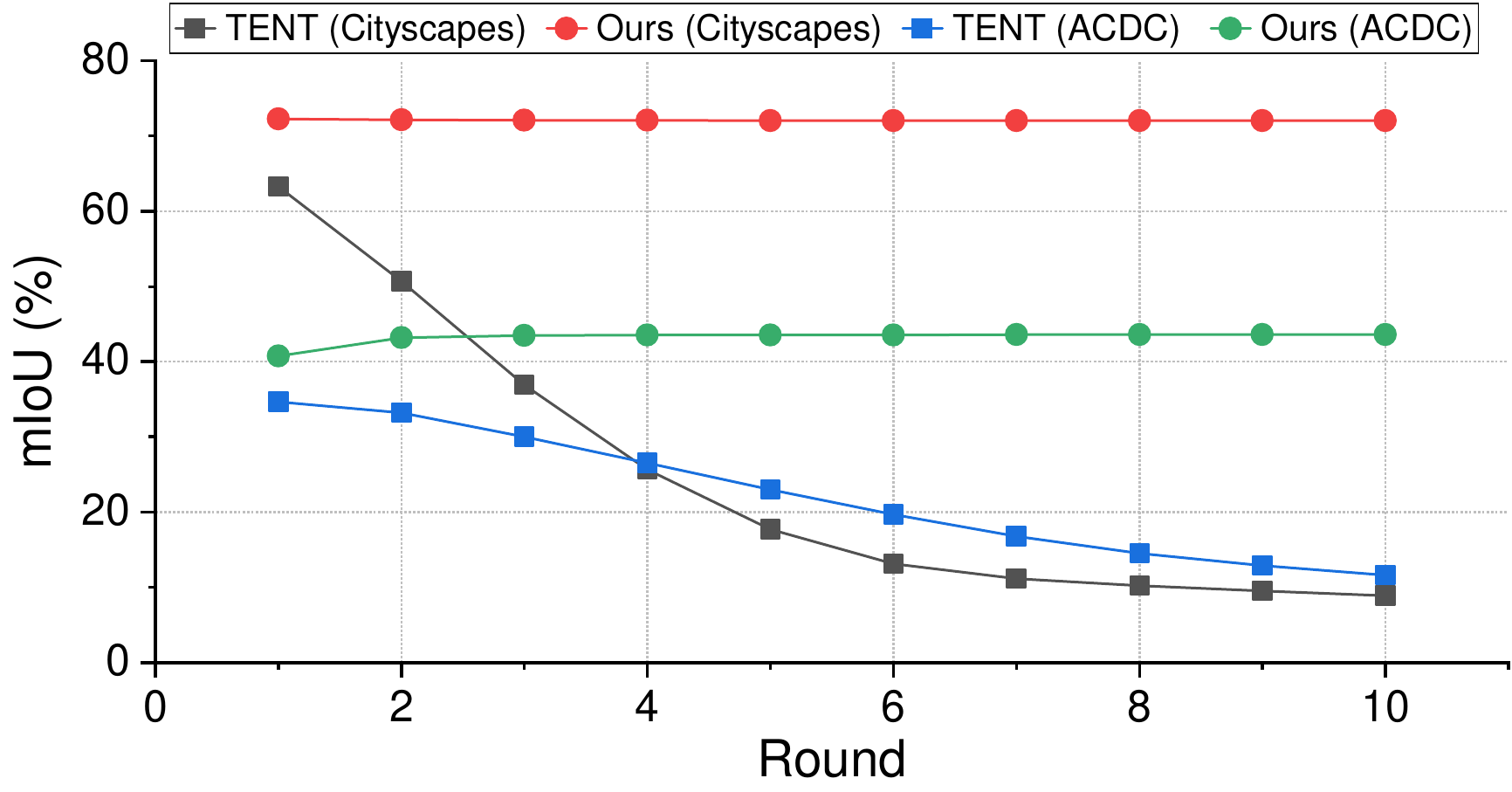}
    \caption{Robustness to catastrophic forgetting. We evaluated the adapted models after each round on the source dataset (Cityscapes). 
    All experiments are evaluated on Cityscapes-to-ACDC in semantic segmentation using DeepLabV3Plus-ResNet50. In comparison to TENT, ours does not show catastrophic forgetting and error accumulation.}
    \label{fig:forgetting}
\end{figure}

\topic{Prevention of catastrophic forgetting.}
As shown in Fig.~\ref{fig:forgetting}, we evaluated the performance of adapted models after each round on the source dataset, to assess the robustness to the forgetting. TENT exhibits an error accumulation, reflected in the decrease in performance on the ACDC with each round. In addition, TENT experiences catastrophic forgetting of source knowledge, as evidenced by a rapid decline in performance on Cityscapes (the source dataset). In contrast, our method demonstrates resilience against forgetting source-trained knowledge, with remarkably improved performance across all rounds.

\subsection{Experiments on Image Classification}
\label{sec:ExpCLF}
\topic{Experimental Setup.} We conduct experiments to verify the effectiveness of our method in image classification. Following CoTTA~\cite{wang2022continual}, we pretrain the WideResNet-28~\cite{zagoruyko2016wide} on CIFAR-10~\cite{krizhevsky2009cifar} and adapt the network to CIFAR-10-C~\cite{hendrycks2018benchmarking} in the single image cTTA setting. 

\topic{Results.} As provided in Table~\ref{tab:CLF_CIFAR10}, our approach outperforms other methods in mean accuracy over different corruption types. TENT shows performance that is almost equal to random performance. The second best method, PETAL, is better than the source model, but it is far from satisfactory performance. In contrast, our method exhibits anti-forgetting and maintains high performance over time.

\section{Conclusion}
In this paper, we propose BESTTA, a style transfer guided continual test-time adaptation (cTTA) method for stable and efficient adaptation, especially in the single image cTTA setting.
To stabilize the adaptation, we devise a stable normalization layer, coined BeIN, that incorporates learnable parameters and source statistics. 
We propose style-guided losses to guide our BeIN to effectively transfer the style of the input image to the source style. 
Our approach achieves state-of-the-art performance and memory efficiency in the single image cTTA setting.
We plan to explore a method to automatically select the best layer to insert our BeIN layer.
{\small    
\bibliographystyle{ieeenat_fullname}
\bibliography{main}
}
{\setlength{\paramarginsize}{2mm} 
    \clearpage
\setcounter{page}{1}
\maketitlesupplementary

\setcounter{section}{0}
\setcounter{table}{0}
\setcounter{figure}{0}
\setcounter{equation}{0}
\renewcommand{\thesection}{\Alph{section}}
\renewcommand{\thetable}{\Alph{table}}
\renewcommand{\thefigure}{\Alph{figure}}
\renewcommand{\theenumi}{\;\;\Alph{enumi}}
\renewcommand{\theequation}{\Alph{equation}}

\section*{Contents}
In this supplementary material, we provide further details of the main paper as follows:
{\setlength{\leftmargin}{2cm}
\begin{enumerate}
    \item Formulation details
    \item Additional ablation studies
    \item Complete implementation details
    \item Further experimental results
\end{enumerate}
}

\hspace{1em}
\hrule
\section{Formulation details}
In this section, we provide the detailed formulations and rationale for our formulation for TTA-as-style-transfer (Eq.~\ref{eq:tta_as_style}) and our proposed BeIN layer presented in Sec.~\ref{sec:probdef} and Sec.~\ref{sec:bein}.
The test-time adaptation task in this section is to adapt a model trained in the source domain that follows $N(\mu_s, \sigma_s^2)$ to the target domain that follows $N(\mu_t, \sigma_t^2)$. Note that the formulations are for a single layer.

\subsection{Problem analysis}
\topic{Instability of batch normalization}
Batch normalization (BN) is formulated as follows:
\begin{equation}
    \text{BN}(X) = \alpha \cdot \frac{X - \mu_{s}}{\sigma_{s}} + \beta
    \label{eq:supp_bn}
\end{equation}
where $\alpha$ and $\beta$ are learnable affine parameters. As shown in Fig.~\ref{fig:supp_prob}, the output of BN layer follows the following distribution:

{\small
\begin{subnumcases}{\text{BN}(X) \sim }
    N(\beta, \alpha^2)  & if  $X \sim  N(\mu_s, \sigma_s^2)$ 
    \label{eq:supp_bn_shift_a}
    \\
    N(\alpha \frac{\mu_t - \mu_s}{\sigma_s} + \beta, \alpha^2 \frac{\sigma_t^2}{\sigma_s^2}) & if  $X \sim N(\mu_t, \sigma_t^2)$ 
    \label{eq:supp_bn_shift_b}
\end{subnumcases}
}
Therefore, the objective of BN is to normalize the input to follow the distribution $N(\beta, \alpha^2)$, however, domain shift causes the distribution misalignment, as shown in Eq. \ref{eq:supp_bn_shift_b}. When $X$ is drawn from the target domain, the difference of distributions makes the BN layers unstable, especially when $\alpha(\mu_t-\mu_s)/\sigma_s$ differs from 0 and $\sigma_t/\sigma_s$ differs from 1.

\topic{TENT}
TENT~\cite{wang2021tent} aligns the output distribution of each BN layers to $N(\beta, \alpha^2)$ as following:
\begin{subequations}
\begin{align}
    \text{BN}_{\text{TENT}}(X) &= {\alpha'} \cdot \frac{X - \mu_X}{\sigma_X} + {\beta'} \label{eq:supp_bn_tent_a} \\
    &\sim N(\alpha' \frac{\mu_t - \mu_X}{\sigma_X} + \beta', \alpha'^2 \frac{\sigma_t^2}{\sigma_X^2}) 
    \label{eq:supp_bn_tent_b}
\end{align}
\end{subequations}
where $\alpha'$ and $\beta'$ are learnable parameters different from $\alpha$ and $\beta$. Due to the distribution of the target domain is unknown, TENT aims to find the optimal values $\alpha'$ and $\beta'$ that approximate the distribution $N(\beta, \alpha^2)$.
The optimal solution to align Eq.~\ref{eq:supp_bn_tent_a} and Eq.~\ref{eq:supp_bn_tent_b} is $\alpha' = \alpha{\sigma_X}/{\sigma_t}$ and $\beta' = \beta - \alpha({\mu_t - \mu_X})/{\sigma_t}$.
Because $\mu_t$ and $\sigma_t$ are unknown, TENT tries to find the optimal parameters by learning with entropy.

\begin{figure}[t!]
    \centering
    \includegraphics[width=\linewidth]{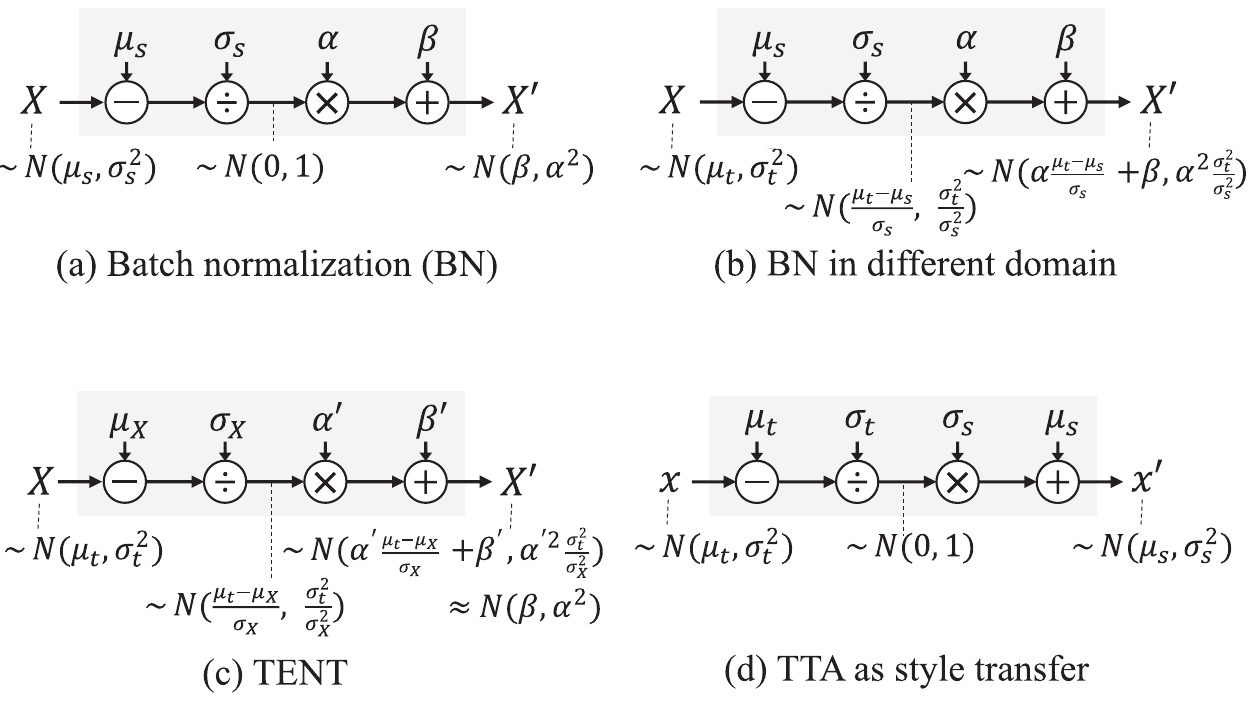}
    \caption{Normalization overview.}
    \label{fig:supp_prob}
\end{figure}

\topic{TTA as style transfer}
We formulate the TTA as a style transfer problem that transfers the target style to the source style (Eq.~\ref{eq:tta_as_style}),
\begin{equation}
    \text{TTA}(x)=\sigma_s \cdot \frac{x - \mu_t}{\sigma_t} + \mu_s
    \label{eq:supp_tta}
\end{equation}
where $x \sim N(\mu_t, \sigma_t^2)$. Then $\text{TTA}(x) \sim N(\mu_s, \sigma_s^2)$, which is aligned to the source distribution.

For example, TENT (Eq.~\ref{eq:tent}) can be viewed as a special case of the TTA-as-style-transfer, whose parameters are $\mu_s = \beta', \sigma_s = \alpha', \mu_t = \mu(X), \sigma_t = \sigma(X)$, and is inserted after every BN layers (i.e., $X = \text{BN}(X_\text{in})$ where $X_\text{in}$ is an input of the BN layer.) In this formulation, TENT is unstable with a small mini-batch because the estimated target statistics $(\mu_t, \sigma_t)$ depends on the unstable sample statistics $(\mu(X), \sigma(X))$.

\vspace{1em}
\subsection{Rationale for the BeIN parameters}
In this section, we provide the rationale for the estimated target statistics $(\hat\mu_t, \hat\sigma_t)$ for BeIN, Eq. \ref{Eq:sigma} and Eq. \ref{Eq:mu}.
As mentioned in Sec.~\ref{sec:bein}, our proposed BeIN is formulated as follows:

\begin{subequations}
\begin{align}
    \text{BeIN}(x) &= \overline{\sigma_s} \cdot \frac{x - \hat\mu_t}{\hat\sigma_t} + \overline{\mu_s} \\
    & \approx TTA(x) \sim N(\overline{\mu_s}, \overline{\sigma_s}^2)
    \label{eq:supp_bein_distrb}
\end{align}
\end{subequations}

\begin{figure}[t!]
    \centering
    \includegraphics[width=\linewidth]{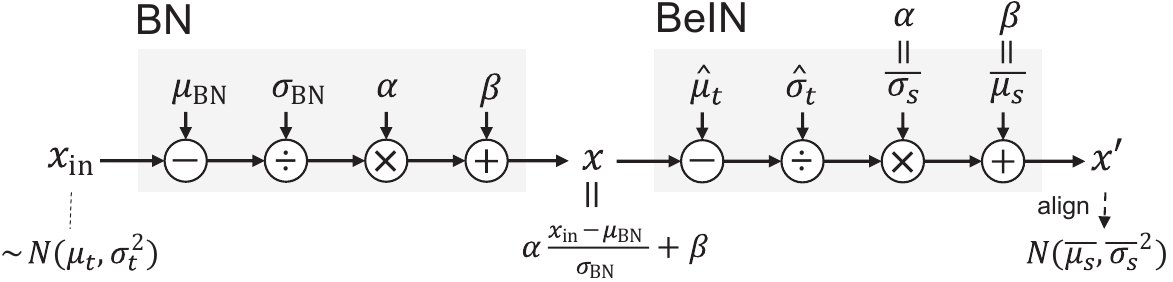}
    \caption{BeIN overview, where BeIN is inserted after a BN layer.}
    \label{fig:supp_bein}
\end{figure}

The objective of BeIN is to align the input distribution to the source distribution $N(\overline{\mu_s}, \overline{\sigma_s}^2)$, as shown in Eq. \ref{eq:supp_bein_distrb}. In our implementation, the objective statistics $\overline{\sigma_s} = \alpha$ and $\overline{\mu_s} = \beta$ since the BeIN layer is inserted after a BN layer as shown in Fig.~\ref{fig:supp_bein}. Let $x_\text{in} \in \mathbb{R}^{C \times HW}$ be the input before the BN layer that follows $N(\mu_t, \sigma_t^2)$. Then the input of the BeIN layer $x = \text{BN}(x_\text{in}) = \alpha ({x_\text{in} - \mu_\text{BN}})/{\sigma_\text{BN}} + \beta$, where $(\mu_\text{BN}, \sigma_\text{BN})$ is the statistics of the source domain on which the BN layer is trained. The ideal BN aligned to the target domain is $\text{BN}_\text{ideal}(x_\text{in}) = \alpha ({x_\text{in} - \mu_t})/{\sigma_t} + \beta$. Consequently, the objective of BeIN is formulated as follows:
\begin{equation}
    \text{BeIN}(x) = \text{BeIN}(\text{BN}(x_\text{in})) = \text{BN}_\text{ideal}(x_\text{in})
    \label{eq:supp_bein_objective}
\end{equation}
A possible solution of Eq.~\ref{eq:supp_bein_objective} is as follows:
\begin{equation}
    \hat\sigma_t = \frac{\sigma_t}{\sigma_\text{BN}} \alpha, \;
    \hat\mu_t = \alpha \frac{\mu_t - \mu_\text{BN}}{\sigma_\text{BN}} + \beta
    \label{eq:supp_bein_sol}
\end{equation}

Since the target statistics $(\mu_t, \sigma_t)$ are unknown, we need to estimate the statistics based on other known statistics. To ensure stability, we estimate the statistics using the stable source statistics as an anchor point, instead of solely relying on the unstable sample statistics.

\topic{Standard deviation}
Let $v = ({x_\text{in} - \mu_t})/{\sigma_t} \sim N(0, 1)$ (i.e., $x_\text{in} = \mu_t + v\sigma_t$.) Then the sample statistics are as follows: 
\begin{equation}
    \sigma_x = \alpha \sigma(v) \frac{\sigma_t}{\sigma_\text{BN}}, \;\;
    \mu_x = \alpha \frac{\mu(v) \sigma_t + \mu_t - \mu_\text{BN}}{\sigma_\text{BN}} + \beta
    \label{eq:supp_bein_sample_stat}
\end{equation}
where $\mu(v)$ and $\sigma(v)$ are the mean and standard variation of $v$, respectively.
By Eq.~\ref{eq:supp_bein_sol} and Eq.~\ref{eq:supp_bein_sample_stat}, the optimal target statistics $(\hat\mu_t, \hat\sigma_t)$ are as follows:
\begin{equation}
    \hat\sigma_t = \frac{\sigma_x}{\sigma(v)}, \;\;
    \hat\mu_t = \mu_x - \mu(v) \hat\sigma_t
    \label{eq:supp_bein_optimal}
\end{equation}

We approximate $\sigma(v) = \sigma_x / \hat\sigma_t$ by the weighted sum of the ideal case ($\sigma(v) = 1$) and the source-sample deviation difference $\sigma_x / \overline{\sigma_s}$ with a learnable parameter $\gamma_\sigma'$:
\begin{equation}
    \sigma(v) \simeq \rho + (1-\rho) \frac{\sigma_x}{\overline{\sigma_s}} + \gamma_\sigma'
    \label{eq:supp_bein_sigma_v}
\end{equation}
Therefore, by Eq. \ref{eq:supp_bein_optimal} and Eq. \ref{eq:supp_bein_sigma_v}, 
\begin{subequations}
\begin{align}
    \frac{1}{\hat\sigma_t} =\ & \frac{\rho}{\sigma_x} + \frac{1-\rho}{\overline{\sigma_s}} + \frac{\gamma_\sigma'}{\sigma_x} \\
    \Longleftrightarrow\;\;&\hat\sigma_t = \frac{\overline{\sigma_s}\cdot\sigma_x}{\rho\overline{\sigma_s} + (1-\rho)\sigma_x + \gamma_{\sigma}}
    \label{eq:supp_bein_sigma}
\end{align}
\end{subequations}
Eq. \ref{eq:supp_bein_sigma} is equivalent to Eq. \ref{Eq:sigma}, where $\gamma_\sigma = \overline{\sigma_s} \gamma_\sigma'$ for simplicity.

\topic{Mean}
Similar to $\sigma(v)$, we approximate $\mu(v) = (\mu_x - \hat\mu_t)/\hat\sigma_t$ by the weighted sum of the ideal case ($\mu(v) = 0$) and the source-sample mean difference $(\mu_x - \overline{\mu_s})/\overline{\mu_s}$ with a learnable parameter $\gamma_\mu'$.

\begin{equation}
    \mu(v) \simeq (1-\rho) \frac{\mu_x - \overline{\mu_s}}{\overline{\sigma_s}} + \gamma_\mu'
    \label{eq:supp_bein_mu_v}
\end{equation}
Therefore, by Eq. \ref{eq:supp_bein_sigma_v} and Eq. \ref{eq:supp_bein_mu_v}, 
\begin{subequations}
\begin{align}
    \hat\mu_t &= \mu_x - \mu(v)\hat\sigma_t \\
    &= \hat\sigma_t (\frac{\mu_x}{\sigma_x} \sigma(v) - \mu(v))  \\
    &\simeq \hat\sigma_t (\rho \frac{\mu_x}{\sigma_x} + (1-\rho) \frac{\overline{\mu_s}}{\overline{\sigma_s}} + \frac{\mu_x}{\sigma_x}\gamma_\sigma' -  \gamma_\mu') \\
    &= \rho \frac{\hat\sigma_t}{\sigma_x}\cdot\mu_x + (1-\rho)\frac{\hat\sigma_t}{\overline{\sigma_s}}\cdot\overline{\mu_s} + \gamma_\mu 
    \label{eq:supp_bein_mu0}
\end{align}
\end{subequations}
Eq. \ref{eq:supp_bein_mu0} is equivalent to Eq. \ref{Eq:mu}, where $\gamma_\mu = \frac{\mu_x}{\sigma_x}\gamma_\sigma' -  \gamma_\mu'$ for simplicity.

\section{Additional ablation studies}
We provide additional ablation studies and the complete results of the ablations studies in Sec.~\ref{sec:ExpSS}.

\begin{table}[t!]
\centering
{\small
\def\arraystretch{1.15}
\setlength{\tabcolsep}{3pt}
\begin{tabularx}{\linewidth}{c|YYY}
\hline
\toprule 
\multirow{2}{*}{Method} & \multicolumn{3}{c}{Batch size}\\
& 1 & 2 & 4  \\
\hline
 Source & 41.8 & 41.8 & 41.8 \\
 TENT-continual~\cite{wang2021tent} & 37.3 & 40.1 & 41.3 \\
 EATA~\cite{niu2022efficient} & 37.4 & 39.2 & 39.6\\
 CoTTA~\cite{wang2022continual} & 42.6 & 42.8 & 42.9 \\
 EcoTTA~\cite{song2023ecotta} &37.4 &38.0 & 38.4 \\
 PETAL~\cite{Brahma2023probabilistic} & 41.8 & 42.0 & 42.0 \\
 \rowcolor{lightgray}
 BESTTA (Ours) & \textbf{44.2} & \textbf{44.2} & \textbf{44.1} \\[-0.3ex] 
\bottomrule
\end{tabularx}
}
\caption{Ablation study on batch sizes. Performances (mIoU in \%) are evaluated on Cityscapes-to-ACDC continual test-time adaptation task, using DeepLabV3Plus-ResNet50~\cite{chen2018deeplabv3+}. Input images are resized to 960$\times$540. Our proposed method consistently achieves the highest performance regardless of batch sizes.}
\label{tab:supp_abalation_batch}
\end{table}
\topic{Effectiveness of batch sizes.} 
We evaluate our method and the baselines on batch sizes of 1, 2, and 4. Table \ref{tab:supp_abalation_batch} shows that our method consistently performs better than the baselines, demonstrating its robustness to batch sizes. 
The BN-based methods (TENT-continual~\cite{wang2021tent}, EATA~\cite{niu2022efficient}, and EcoTTA~\cite{song2023ecotta}) improve with larger batch sizes due to reduced instability, whereas CoTTA~\cite{wang2022continual} and PETAL~\cite{Brahma2023probabilistic} do not exhibit significant improvement.
Despite their improved performance, our method consistently surpasses the baselines regardless of the batch sizes.

\begin{table*}[t!]
\centering
\small
\def\arraystretch{1.15}
\setlength{\tabcolsep}{3pt}
\def\tabmid{\centering\arraybackslash}
\def\losslsp{1.5cm}
\def\lossrsp{1.4cm}
\begin{tabularx}{\linewidth}{>{\centering\arraybackslash}m{0.65cm}|YYYY|YYYY|Y}
\hline
\toprule
& $\mathcal{L}_{entropy}$ & $\mathcal{L}_{style}$ & $\mathcal{L}_{content}$ & $\mathcal{L}_{L2}$ & {Round 1} & {Round 4} & {Round 7} & {Round 10} & {Mean}\\
\hline 
(a) & \checkmark & \xmark & \xmark & \xmark  & 21.2 & 5.3 & 4.5 & 4.2 & 6.7 \\
(b) & \xmark & \checkmark & \xmark & \xmark & 	31.0 & 25.4 & 24.4 & 24.0 & 25.6 \\
\hline
(c) & \checkmark & \checkmark& \xmark & \xmark  & 37.1 & 32.3 & 30.2 & 28.9 & 31.8  \\
(d) & \checkmark & \xmark & \checkmark & \xmark & 		21.8 & 14.0 & 13.6 & 13.4 & 14.7 \\
(e) & \checkmark & \xmark & \xmark & \checkmark & 13.8 & 5.0 & 5.0 & 5.0 & 5.9 \\	
(f) & \xmark & \checkmark & \checkmark & \xmark & 	39.0 & 37.3 & 37.1 & 36.9 & 37.4 \\
(g) & \xmark & \checkmark & \xmark & \checkmark & 		38.3 & 38.6 & 38.5 & 38.6 & 38.5 \\
\hline
(h) & \checkmark & \checkmark & \checkmark & \xmark  & 40.1 & 39.2 & 37.9 & 37.3 & 38.6 \\
(i) & \checkmark & \checkmark & \xmark & \checkmark & 		28.8 & 26.9 & 26.9 & 26.9 & 27.1 \\
(j) & \checkmark & \xmark & \checkmark & \checkmark & 		37.2 & 41.1 & 41.1 & 41.1 & 40.7 \\
(k) & \xmark & \checkmark & \checkmark & \checkmark & 	40.4 & 40.7 & 40.7 & 40.8 & 40.7 \\
\hline
\rowcolor{lightgray}
(l) & \checkmark & \checkmark & \checkmark & \checkmark  & \textbf{40.8} & \textbf{43.5} & \textbf{43.6} & \textbf{43.6} & \textbf{43.2} \\[-0.3ex]
\bottomrule
\end{tabularx}
\caption{Ablation study on our losses. Performances (mIoU in \%) are evaluated on Cityscapes-to-ACDC single image continual test-time adaptation task for each different combination of losses, using DeepLabV3Plus-ResNet50~\cite{chen2018deeplabv3+}. The results are separated by the number of losses used. Using all of our proposed losses considerably improves the performance. Because the content loss and the L2 regularization are only meaningful with other losses, their individual results are excluded.}
\label{tab:supp_ablation_loss}

\end{table*}
\topic{Effectiveness of losses.} 
Table~\ref{tab:supp_ablation_loss} displays the complete results of the ablation study on losses in Table~\ref{tab:ablation_loss}. The results indicate that the proposed losses presented in Sec.~\ref{sec:loss} effectively enables the single-image continual test-time adaptation as follows. Firstly, the style loss outperforms the entropy loss, confirming its effectiveness. Secondly, using both the style loss and the entropy enhances the performance, indicating the stability of the style loss. Thirdly, the use of content loss consistently improves the performance, indicating the effectiveness of the content loss. Lastly, the L2 regularization loss helps performance by preventing the catastrophic forgetting, which is especially effective when used with the style loss and the content loss.


\topic{Layer selection to insert BeIN Layer.}
Table~\ref{tab:supp_abalation_layer2} provides the complete results of the ablation study on the layer selection to insert the proposed BeIN layer in Table~\ref{tab:abalation_layer}. 
The results indicate that inserting the BeIN layer after \textit{Layer3} consistently performs better than other cases in terms of mIoU in each round, regardless of the number of BeIN layers used.
Additionally, inserting a BeIN layer after \textit{Layer1} achieves better adaptation performance in fog and snow condition, as presented in (e), (f), and (m). This result indicates that selecting a layer in the lower level is advantageous in the presence of prevalent adverse weather effects, as the low-level features capture the prevalent patterns~\cite{fahes2023poda}. The results suggest that inserting a BeIN layer following a higher-level layer, excluding the last layer, is more effective. This is due to the ability of the higher-level layers to capture the global adverse weather effects such as rain accumulation.

Notably, using a single BeIN layer after \textit{Layer3} achieves the highest performance, as demonstrated in (c). This implies that \textit{Layer3} captures both low- and high-level features, enabling effective style transfer guided adaptation. However, using more BeIN layers does not improve performance, as shown in (e-o). We speculate that the normalization process becomes unstable when transferring the style more than once.
\begin{table*}[t!]
\centering
\small
\def\arraystretch{1.15}
\setlength{\tabcolsep}{0.5pt}
\begin{tabularx}{\linewidth}{>{\centering\arraybackslash}m{0.65cm}|YYYY|YYYY|YYYY|Y}
\hline
\toprule
 & \textit{Layer1} & \textit{Layer2} & \textit{Layer3} & \textit{Layer4} & Fog & Night & Rain & Snow & Round 1 & Round 4 & Round 7 & Round 10 & Mean \\
\hline
(a) & \checkmark & \xmark & \xmark & \xmark & 	53.6 & 22.2 & 46.5 & 44.4 & 	\underline{40.2} & 42.0 & 41.9 & 41.9 & 41.7 \\
(b) & \xmark & \checkmark & \xmark & \xmark & 	52.6 & 21.8 & 45.9 & 43.4 & 	39.3 & 41.1 & 41.1 & 41.1 & 40.9 \\
\rowcolor{lightgray}
(c) & \xmark & \xmark & \checkmark & \xmark & 	\underline{53.7} & \underline{26.0} & \textbf{48.3} & 45.0 & 	\textbf{40.8} & \textbf{43.5} & \textbf{43.6} & \textbf{43.6} & \textbf{43.2} \\
(d) & \xmark & \xmark & \xmark & \checkmark & 	42.1 & 23.5 & 40.2 & 38.1 & 	36.0 & 36.0 & 36.0 & 36.0 & 36.0 \\
\hline
(e) & \checkmark & \checkmark & \xmark & \xmark & 	54.2 & 17.8 & 47.5 & 44.9 & 	39.7 & 41.2 & 41.2 & 41.2 & 41.1 \\
(f) & \checkmark & \xmark & \checkmark & \xmark & 	\textbf{54.3} & 23.3 & 48.2 & \underline{46.2} & 	38.7 & \textbf{43.5} & \underline{43.5} & \underline{43.5} & \underline{43.0} \\
(g) & \checkmark & \xmark & \xmark & \checkmark & 	47.7 & \textbf{26.1} & 45.2 & 42.6 & 	40.1 & 40.4 & 40.4 & 40.4 & 40.4 \\
(h) & \xmark & \checkmark & \checkmark & \xmark & 	52.5 & 23.2 & 46.5 & 44.7 & 	39.6 & 42.0 & 41.9 & 41.9 & 41.7 \\
(i) & \xmark & \checkmark & \xmark & \checkmark & 	49.6 & 25.3 & 44.3 & 41.9 & 	39.1 & 40.4 & 40.4 & 40.4 & 40.3 \\
(j) & \xmark & \xmark & \checkmark & \checkmark & 	51.9 & 25.0 & 45.8 & 42.8 & 	39.0 & 41.6 & 41.6 & 41.6 & 41.4 \\
\hline
(k) & \checkmark & \checkmark & \checkmark & \xmark & 	53.0 & 23.0 & 47.0 & 45.4 & 	38.4 & 42.5 & 42.5 & 42.5 & 42.1 \\
(l) & \checkmark & \checkmark & \xmark & \checkmark & 	51.2 & 21.0 & 45.4 & 43.7 & 	39.2 & 40.5 & 40.4 & 40.4 & 40.3 \\
(m) & \checkmark & \xmark & \checkmark & \checkmark & 	53.2 & 25.3 & \underline{47.3} & \textbf{46.3} & 	40.0 & \underline{43.4} & 43.3 & 43.4 & \underline{43.0} \\
(n) & \xmark & \checkmark & \checkmark & \checkmark & 	51.3 & 23.5 & 45.5 & 43.9 & 	38.1 & 41.4 & 41.4 & 41.4 & 41.0 \\
\hline
(o) & \checkmark & \checkmark & \checkmark & \checkmark & 	52.0 & 23.6 & 46.5 & 45.2 & 	38.5 & 42.2 & 42.1 & 42.2 & 41.8 \\

\bottomrule
\end{tabularx}
\caption{Ablation study on the position of BeIN. Performances (mIoU in \%) are evaluated on Cityscapes-to-ACDC single image continual test-time adaptation task, using DeepLabV3Plus-ResNet50~\cite{chen2018deeplabv3+}. The results are separated by the number of layers where BeIN is inserted.}
\label{tab:supp_abalation_layer2}
\end{table*}

{\setlength{\paramarginsize}{1.0mm} 
\section{Complete implementation details}
We provide further implementation details for our experiments in Sec.~\ref{sec:exp}.

\topic{Details for the correlations}
To examine the correlations between adaptation performance and style transfer related metrics in Fig.~\ref{fig:loss_correlation} and Fig.~\ref{fig:supp_loss_clf}, we measure direction similarity, target similarity, entropy, and source similarity, which are described in Sec.~\ref{sec:loss}. We measure the similarities between adapted and unadapted embeddings, using the method suggested by Schneider \textit{et al.}~\cite{schneider2020improving}, a simple unsupervised adaptation method that employs the weighted mean of batch statistics derived from both the source and target data. We set the hyperparameter $N$ to 8, as suggested by the authors, where $N$ determines the weights for the source and target statistics. The colored regions in Fig.~\ref{fig:loss_correlation} and Fig.~\ref{fig:supp_loss_clf} indicate the 95\% confidence interval.

\begin{figure}[t!]
    \centering
    \includegraphics[width=\linewidth]{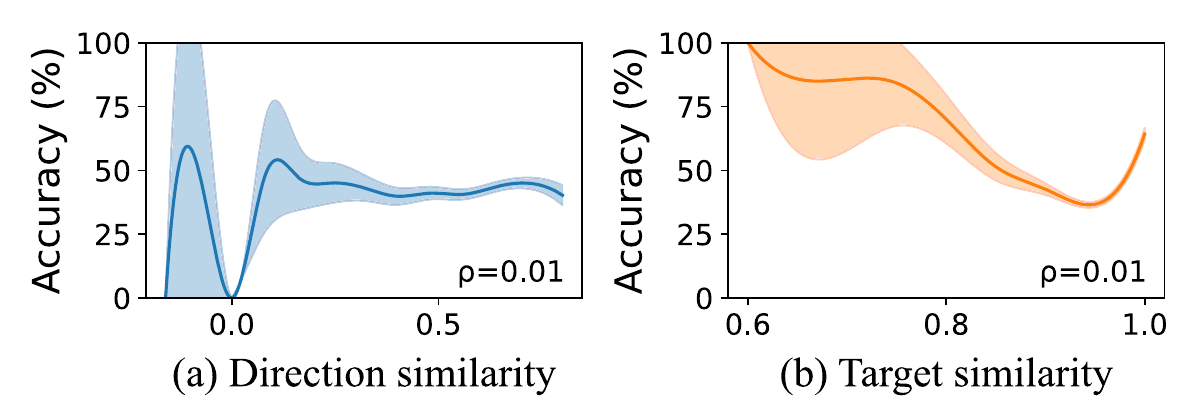}
    \caption{Correlations between performance and style transfer related metrics in image classification. In contrast to semantic segmentation, we discover that direction and target similarities are uncorrelated ($\rho < 0.05$). Target similarity appears to be related in (b), however, the majority of data (93.7\%) have target similarity above 0.9, making the results unreliable.  
    All results are evaluated on CIFAR-10-C~\cite{hendrycks2018benchmarking} using WideResNet-28~\cite{zagoruyko2016wide} pretrained on CIFAR-10~\cite{krizhevsky2009cifar}.}
    \label{fig:supp_loss_clf}
\end{figure}

\topic{Details for image classification}
In the image classification experiments in Sec.~\ref{sec:ExpCLF}, we set the learning rate to 0.1, $\rho$ to 0.9 and $\lambda_1$, $\lambda_2$, $\lambda_3$, and $\lambda_4$ to 1, 0, 1, and 0, respectively. 
Based on our empirical observations depicted in Fig.~\ref{fig:supp_loss_clf}, the content loss is not effective in image classification, thus, we do not utilize it in image classification. This is because the image classification data are synthetic, the content of the data remains unchanged in the presence of image corruption. We replace the style loss with the cosine distance between $\overline{\mathcal{E}(x_{src})}$ and $\mathcal{E}'(x)$, i.e. $\mathcal{L}_{style}=1-cos(\overline{\mathcal{E}(x_{src})}, \mathcal{E}'(x))$. We find that employing this straightforward form of style loss is more effective in guiding the BeIN layer to successfully transfer the style of the input image in image classification.

\topic{Other details}
In Sec.~\ref{sec:ExpSS}, we set $\rho$ to 0.7 and set the momentum value of the optimizer to 0.9. For baselines, we use the optimal configurations provided in their respective papers. We measure the peak GPU memory consumption using 
\texttt{torch.cuda.max\char`_memory\char`_allocated()}
method. We measure the average processing time per image by averaging the wall time elapsed for processing 1,000 images. The average processing time includes both the forward and backward operations.


%

\section{Further experimental results}
We present the further experimental results including the complete results of the experiments in Sec.~\ref{sec:exp}.

\topic{Complete results} 
Table~\ref{tab:supp_ACDC} presents the complete results of Table~\ref{tab:SS_ACDC}. Our method consistently outperforms the baselines in all rounds regardless of the weather conditions on the Cityscapes-to-ACDC single image continual test-time adaptation task. Our method consumes the lowest GPU memory and the second lowest processing time among the backward-based baseline test-time adaptation methods.
Table~\ref{tab:supp_CSC} shows the complete results of Table~\ref{tab:main-cityscapes-c-gradual}. Our method achieves the highest mean mIoU regardless of the type of corruptions on the Cityscapes-to-Cityscapes-C single image gradual test-time adaptation task.
\begin{table*}[t!]
{\small
\begin{subtable}[h]{\textwidth}
    \centering
    \def\arraystretch{1.2}
    \setlength{\tabcolsep}{0.5pt}
    \begin{tabularx}{\textwidth}{c|*{3}{YYYY>{\columncolor{lightgray}}Y|}YYYY>{\columncolor{lightgray}}Y}
    \hline
    \toprule 
     Time & \multicolumn{20}{l}{$\; t \; \xrightarrow{\hspace{0.8\textwidth}}$}  \\
    \hline
     \multirow{2}{*}{Method} 
     & \multicolumn{5}{c|}{Round 1} &  \multicolumn{5}{c|}{Round 2} &  \multicolumn{5}{c|}{Round 3} &  \multicolumn{5}{c}{Round 4} \\
     & Fog & Night & Rain & Snow & Mean & Fog & Night & Rain & Snow & Mean & Fog & Night & Rain & Snow & Mean & Fog & Night & Rain & Snow & Mean \\
     \hline
    Source & 44.3 & 22.0 & 41.5 & 38.9 & 36.6 & 44.3 & 22.0 & 41.5 & 38.9 & 36.6 & 44.3 & 22.0 & 41.5 & 38.9 & 36.6 & 44.3 & 22.0 & 41.5 & 38.9 & 36.6 \\
    BN Stats Adapt~\cite{nado2021evaluating} & 36.8 & 23.5 & 38.2 & 36.3 & 33.7 & 36.8 & \underline{23.5} & 38.2 & 36.3 & 33.7 & 36.8 & \underline{23.5} & 38.2 & 36.3 & 33.7 & 36.8 & 23.5 & 38.2 & 36.3 & 33.7 \\
    \hline
    TENT*~\cite{wang2021tent} & 38.2 & 22.9 & 41.1 & 37.8 & 35.0 & 38.2 & 22.9 & 41.1 & 37.8 & 35.0 & 38.2 & 22.9 & 41.1 & 37.8 & 35.0 & 38.2 & 22.9 & 41.1 & 37.8 & 35.0 \\
    TENT-continual~\cite{wang2021tent} & 37.7 & \underline{23.5} & 39.9 & 37.5 & 34.6 & 39.0 & 22.1 & 37.6 & 34.0 & 33.2 & 35.5 & 19.9 & 34.3 & 30.3 & 30.0 & 31.4 & 17.3 & 30.7 & 26.8 & 26.6 \\
    EATA~\cite{niu2022efficient} & 36.8 & 23.4 & 38.3 & 36.5 & 33.7 & 37.1 & 23.4 & 38.5 & 36.7 & 33.9 & 37.3 & 23.5 & 38.8 & 36.9 & 34.1 & 37.6 & \underline{23.5} & 39.0 & 37.1 & 34.3 \\
    CoTTA~\cite{wang2022continual} & \underline{46.3} & 22.0 & \underline{44.2} & \underline{40.4} & \underline{38.2} & \underline{47.3} & 21.7 & \underline{45.0} & 40.4 & \underline{38.6} & \underline{47.8} & 21.4 & \underline{45.3} & 40.3 & \underline{38.7} & \underline{48.1} & 21.0 & \underline{45.3} & 40.2 & 38.6 \\
    EcoTTA~\cite{song2023ecotta} & 33.4 & 20.8 & 35.6 & 32.9 & 30.7 & 33.7 & 21.0 & 35.8 & 33.0 & 30.9 & 34.0 & 21.1 & 35.9 & 33.1 & 31.0 & 34.2 & 21.2 & 36.1 & 33.2 & 31.1 \\
    PETAL~\cite{Brahma2023probabilistic} & 44.7 & 22.1 & 42.9 & 40.1 & 37.5 & 46.6 & 22.4 & 43.8 & \underline{40.6} & 38.4 & 47.2 & 22.4 & 44.2 & \underline{40.7} & 38.6 & 47.3 & 22.4 & 44.4 & \underline{40.8} & \underline{38.7} \\
    BESTTA (Ours) & \textbf{47.8} & \textbf{24.3} & \textbf{47.2} & \textbf{43.8} & \textbf{40.8} & \textbf{53.4} & \textbf{26.3} & \textbf{48.3} & \textbf{44.8} & \textbf{43.2} & \textbf{54.3} & \textbf{26.2} & \textbf{48.4} & \textbf{45.1} & \textbf{43.5} & \textbf{54.5} & \textbf{26.1} & \textbf{48.4} & \textbf{45.2} & \textbf{43.5} \\[-0.3ex]
    \bottomrule
    \end{tabularx}
   \caption{\small Round 1--4 \vspace{2em}}
   \label{tab:supp_ACDC_r1_4}
\end{subtable}

\begin{subtable}[h]{\textwidth}
    \centering
    \def\arraystretch{1.2}
    \setlength{\tabcolsep}{0.5pt}
    \begin{tabularx}{\textwidth}{c|*{3}{YYYY>{\columncolor{lightgray}}Y|}YYYY>{\columncolor{lightgray}}Y}
    \hline
    \toprule 
     Time & \multicolumn{20}{l}{$\; t \; \xrightarrow{\hspace{0.8\textwidth}}$}  \\
    \hline
     \multirow{2}{*}{Method} 
     & \multicolumn{5}{c|}{Round 5} &  \multicolumn{5}{c|}{Round 6} &  \multicolumn{5}{c|}{Round 7} &  \multicolumn{5}{c}{Round 8} \\
     & Fog & Night & Rain & Snow & Mean & Fog & Night & Rain & Snow & Mean & Fog & Night & Rain & Snow & Mean & Fog & Night & Rain & Snow & Mean \\
     \hline
    Source & 44.3 & 22.0 & 41.5 & 38.9 & 36.6 & 44.3 & 22.0 & 41.5 & 38.9 & 36.6 & 44.3 & 22.0 & 41.5 & 38.9 & 36.6 & 44.3 & 22.0 & 41.5 & 38.9 & 36.6 \\
    BN Stats Adapt~\cite{nado2021evaluating} & 36.8 & 23.5 & 38.2 & 36.3 & 33.7 & 36.8 & 23.5 & 38.2 & 36.3 & 33.7 & 36.8 & 23.5 & 38.2 & 36.3 & 33.7 & 36.8 & 23.5 & 38.2 & 36.3 & 33.7 \\
    \hline
    TENT*~\cite{wang2021tent} & 38.2 & 22.9 & 41.1 & 37.8 & 35.0 & 38.2 & 22.9 & 41.1 & 37.8 & 35.0 & 38.2 & 22.9 & 41.1 & 37.8 & 35.0 & 38.2 & 22.9 & 41.1 & 37.8 & 35.0 \\
    TENT-continual~\cite{wang2021tent} & 27.2 & 14.7 & 26.6 & 23.4 & 23.0 & 23.1 & 12.7 & 22.6 & 20.2 & 19.7 & 19.8 & 11.2 & 18.8 & 17.3 & 16.8 & 17.2 & 10.1 & 15.9 & 15.0 & 14.5 \\
    EATA~\cite{niu2022efficient} & 37.7 & \underline{23.5} & 39.2 & 37.3 & 34.4 & 37.9 & \underline{23.5} & 39.4 & 37.5 & 34.6 & 38.1 & \underline{23.6} & 39.5 & 37.6 & 34.7 & 38.2 & \underline{23.6} & 39.7 & 37.8 & 34.8 \\
    CoTTA~\cite{wang2022continual} & \underline{48.1} & 20.8 & \underline{45.4} & 40.0 & 38.6 & \underline{48.1} & 20.5 & \underline{45.3} & 39.9 & 38.5 & \underline{48.1} & 20.4 & \underline{45.3} & 39.7 & 38.4 & \underline{48.1} & 20.2 & \underline{45.2} & 39.6 & 38.3 \\
    EcoTTA~\cite{song2023ecotta} & 34.4 & 21.2 & 36.2 & 33.2 & 31.3 & 34.5 & 21.3 & 36.3 & 33.2 & 31.3 & 34.7 & 21.3 & 36.3 & 33.2 & 31.4 & 34.8 & 21.4 & 36.4 & 33.2 & 31.4 \\
    PETAL~\cite{Brahma2023probabilistic} & 47.3 & 22.3 & 44.5 & \underline{40.7} & \underline{38.7} & 47.3 & 22.2 & 44.6 & \underline{40.6} & \underline{38.7} & 47.1 & 22.1 & 44.7 & \underline{40.6} & \underline{38.6} & 47.1 & 22.0 & 44.6 & \underline{40.6} & \underline{38.6} \\
    BESTTA (Ours) & \textbf{54.5} & \textbf{26.1} & \textbf{48.4} & \textbf{45.2} & \textbf{43.6} & \textbf{54.5} & \textbf{26.1} & \textbf{48.4} & \textbf{45.3} & \textbf{43.6} & \textbf{54.5} & \textbf{26.1} & \textbf{48.4} & \textbf{45.3} & \textbf{43.6} & \textbf{54.5} & \textbf{26.1} & \textbf{48.4} & \textbf{45.3} & \textbf{43.6} \\[-0.3ex]
    \bottomrule
    \end{tabularx}
   \caption{\small Round 5-8 \vspace{2em}}
   \label{tab:supp_ACDC_r5_8}
\end{subtable}

\begin{subtable}[h]{\textwidth}
    \centering
    \def\arraystretch{1.2}
    \setlength{\tabcolsep}{1pt}
    \begin{tabularx}{\textwidth}{c|
    *{2}{YYYY>{\columncolor{lightgray}}Y|}
    *{2}{>{\columncolor{lightgray}}Y}
    >{\columncolor{lightgray}}c}
    \hline
    \toprule 
     Time & \multicolumn{10}{l|}{$\; t \; \xrightarrow{\hspace{0.59\textwidth}}$} & \multirow{3}{*}{} &&  \\
    \cline{1-11}
     \multirow{2}{*}{Method} 
     & \multicolumn{5}{c|}{Round 9} &  \multicolumn{5}{c|}{Round 10} &&&  \\
     & Fog & Night & Rain & Snow & Mean & Fog & Night & Rain & Snow & Mean & \multirow{-3}{*}{\shortstack{Mean $\uparrow$\\(\%)}} & \multirow{-3}{*}{\shortstack{Time $\downarrow$\\(ms)}} & \multirow{-3}{*}{\shortstack{Memory $\downarrow$\\(GB)}} \\
     \hline
    Source & 44.3 & 22.0 & 41.5 & 38.9 & 36.6 & 44.3 & 22.0 & 41.5 & 38.9 & 36.6 & 36.6 & 135.9 & 1.76 \\
    BN Stats Adapt~\cite{nado2021evaluating} & 36.8 & 23.5 & 38.2 & 36.3 & 33.7 & 36.8 & 23.5 & 38.2 & 36.3 & 33.7 & 33.7 & 192.7 & 2.01 \\
    \hline
    TENT*~\cite{wang2021tent} & 38.2 & 22.9 & 41.1 & 37.8 & 35.0 & 38.2 & 22.9 & 41.1 & 37.8 & 35.0 & 35.0 & 343.8 & \underline{9.58} \\
    TENT-continual~\cite{wang2021tent} & 15.1 & 9.3 & 13.9 & 13.2 & 12.9 & 13.4 & 8.7 & 12.4 & 11.8 & 11.6 & 22.3 & 343.8 & \underline{9.58} \\
    EATA~\cite{niu2022efficient} & 38.3 & \underline{23.6} & 39.8 & 37.9 & 34.9 & 38.5 & \underline{23.6} & 39.9 & 38.0 & 35.0 & 34.4 & \textbf{190.6} & 9.90 \\
    CoTTA~\cite{wang2022continual} & \underline{48.0} & 20.1 & \underline{45.2} & 39.6 & 38.2 & \underline{48.0} & 20.0 & \underline{45.1} & 39.5 & 38.1 & 38.4 & 4337 & 18.20 \\
    EcoTTA~\cite{song2023ecotta} & 34.9 & 21.4 & 36.4 & 33.2 & 31.5 & 34.9 & 21.4 & 36.5 & 33.2 & 31.5 & 31.2 & 759.3 & 10.90 \\
    PETAL~\cite{Brahma2023probabilistic} & 47.0 & 21.9 & 44.6 & \underline{40.5} & \underline{38.5} & 46.9 & 22.0 & 44.6 & \underline{40.5} & \underline{38.5} & \underline{38.5} & 5120 & 18.51 \\
    BESTTA (Ours) & \textbf{54.5} & \textbf{26.1} & \textbf{48.4} & \textbf{45.3} & \textbf{43.6} & \textbf{54.5} & \textbf{26.1} & \textbf{48.4} & \textbf{45.3} & \textbf{43.6} & \textbf{43.2} & \underline{291.8} & \textbf{8.77} \\[-0.3ex]
    \bottomrule
    \end{tabularx}
   \caption{\small Round 9-10}
   \label{tab:supp_ACDC_r9_10}
\end{subtable}
}
\caption{Semantic segmentation results (mIoU in \%) on Cityscapes-to-ACDC single image continual test-time adaptation task. We compare ours with other state-of-the-art TTA methods in terms of peak GPU memory usage (GB) and time consumption (ms) for each iteration, and mean intersection over union (mIoU). All results are evaluated using the DeepLabV3Plus-ResNet50~\cite{chen2018deeplabv3+}. $^*$ denotes the requirement about when the domain shift occurs. The \textbf{best} and \underline{second best} results among the backward-based methods are highlighted. \vspace{1em}}
\label{tab:supp_ACDC}
\end{table*}
\begin{table*}[t!]
{\small
\def\timearrowwidth{0.68\textwidth}
\def\vspacesubtables{\vspace{2em}}
\def\arraystretches{1.2}
\vspace{2em}
\begin{subtable}[t]{\textwidth}
    \centering
    \def\arraystretch{\arraystretches}
    \setlength{\tabcolsep}{5pt}
    \begin{tabularx}{\textwidth}{c|*{9}{Y}|>{\columncolor{lightgray}}Y}
    \hline 
    \toprule 
     Time & \multicolumn{9}{l|}{$\; t \; \xrightarrow{\hspace{\timearrowwidth}}$} &  \\
    \cline{1-10}
     Severity & 1 & 2 & 3 & 4 & 5 & 4 & 3 & 2 & 1 & \multirow{-2}{*}{Mean} \\
     \hline
    Source & \textbf{73.2} & \textbf{71.1} & \underline{67.7} & \underline{63.1} & 57.2 & \underline{63.1} & \underline{67.7} & \textbf{71.1} & \textbf{73.2} & \underline{67.5} \\
    TENT-continual~\cite{wang2021tent} & 70.0 & 67.4 & 64.9 & 62.1 & \underline{58.8} & 59.1 & 58.8 & 58.2 & 57.5 & 61.9 \\
    PETAL~\cite{Brahma2023probabilistic} & \underline{73.1} & \underline{70.8} & 67.0 & 61.6 & 54.2 & 59.1 & 63.0 & 65.9 & 68.2 & 64.7 \\
    BESTTA (Ours) & 72.9 & 70.7 & \textbf{68.3} & \textbf{65.3} & \textbf{61.4} & \textbf{65.0} & \textbf{67.8} & \underline{70.1} & \underline{71.9} & \textbf{68.2} \\[-0.3ex]
    \bottomrule
    \end{tabularx}
   \caption{\small Brightness \vspacesubtables}
   \label{tab:supp_CSC_r1}
\end{subtable}

\begin{subtable}[h]{\textwidth}
    \centering
    \def\arraystretch{\arraystretches}
    \setlength{\tabcolsep}{5pt}
    \begin{tabularx}{\textwidth}{c|*{9}{Y}|>{\columncolor{lightgray}}Y}
    \hline
    \toprule 
     Time & \multicolumn{9}{l|}{$\; t \; \xrightarrow{\hspace{\timearrowwidth}}$} &  \\
    \cline{1-10}
     Severity & 1 & 2 & 3 & 4 & 5 & 4 & 3 & 2 & 1 & \multirow{-2}{*}{Mean} \\
     \hline
    Source & \textbf{67.1} & \underline{63.5} & \underline{58.0} & \underline{55.7} & \underline{44.8} & \underline{55.7} & \underline{58.0} & \underline{63.5} & \textbf{67.1} & \underline{59.3} \\
    TENT-continual~\cite{wang2021tent} & 53.5 & 50.9 & 48.2 & 45.3 & 40.7 & 42.0 & 42.2 & 42.1 & 42.0 & 45.2 \\
    PETAL~\cite{Brahma2023probabilistic} & 58.0 & 50.7 & 41.3 & 32.5 & 17.8 & 25.6 & 28.7 & 31.9 & 36.2 & 35.8 \\
    BESTTA (Ours) & \underline{66.5} & \textbf{64.0} & \textbf{59.6} & \textbf{57.6} & \textbf{50.4} & \textbf{57.2} & \textbf{59.3} & \textbf{63.6} & \underline{66.1} & \textbf{60.5} \\[-0.3ex]
    \bottomrule
    \end{tabularx}
   \caption{\small Fog \vspacesubtables}
   \label{tab:supp_CSC_r2}
\end{subtable}

\begin{subtable}[h]{\textwidth}
    \centering
    \def\arraystretch{\arraystretches}
    \setlength{\tabcolsep}{5pt}
    \begin{tabularx}{\textwidth}{c|*{9}{Y}|>{\columncolor{lightgray}}Y}
    \hline
    \toprule 
     Time & \multicolumn{9}{l|}{$\; t \; \xrightarrow{\hspace{\timearrowwidth}}$} &  \\
    \cline{1-10}
     Severity & 1 & 2 & 3 & 4 & 5 & 4 & 3 & 2 & 1 & \multirow{-2}{*}{Mean} \\
     \hline
    Source & \underline{44.9} & 25.6 & 18.0 & 16.3 & 13.5 & 16.3 & 18.0 & \underline{25.6} & \underline{44.9} & \underline{24.8} \\
    TENT-continual~\cite{wang2021tent} & 33.8 & \underline{27.7} & \underline{23.5} & \underline{21.6} & \textbf{19.6} & \underline{19.5} & \underline{19.3} & 20.1 & 21.6 & 23.0 \\
    PETAL~\cite{Brahma2023probabilistic} & 17.7 & 6.6 & 4.5 & 3.6 & 3.1 & 3.0 & 3.2 & 3.3 & 6.0 & 5.7 \\
    BESTTA (Ours) & \textbf{53.8} & \textbf{38.4} & \textbf{28.1} & \textbf{24.1} & \underline{17.9} & \textbf{20.5} & \textbf{21.9} & \textbf{30.1} & \textbf{49.6} & \textbf{31.6} \\[-0.3ex]
    \bottomrule
    \end{tabularx}
   \caption{\small Frost \vspacesubtables}
   \label{tab:supp_CSC_r3}
\end{subtable}

\begin{subtable}[h]{\textwidth}
    \centering
    \def\arraystretch{\arraystretches}
    \setlength{\tabcolsep}{5pt}
    \begin{tabularx}{\textwidth}{c|*{9}{Y}|>{\columncolor{lightgray}}Y}
    \hline
    \toprule 
     Time & \multicolumn{9}{l|}{$\; t \; \xrightarrow{\hspace{\timearrowwidth}}$} & \\
    \cline{1-10}
     Severity & 1 & 2 & 3 & 4 & 5 & 4 & 3 & 2 & 1 & \multirow{-2}{*}{Mean} \\
     \hline
    Source & \underline{25.3} & 11.8 & 11.5 & 9.1 & 9.5 & 9.1 & 11.5 & 11.8 & \underline{25.3} & 13.9 \\
    TENT-continual~\cite{wang2021tent} & 19.3 & \underline{16.8} & \underline{16.7} & \textbf{15.1} & \textbf{14.2} & \textbf{13.9} & \underline{14.0} & \underline{13.2} & 14.2 & \underline{15.3}\\
    PETAL~\cite{Brahma2023probabilistic} & 0.7 & 0.5 & 0.3 & 0.3 & 0.5 & 0.3 & 0.3 & 0.3 & 0.4 & 0.4\\
    BESTTA (Ours) & \textbf{39.5} & \textbf{20.4} & \textbf{18.6} & \underline{12.8} & \underline{12.5} & \underline{12.2} & \textbf{16.0} & \textbf{16.9} & \textbf{30.7} & \textbf{20.0} \\[-0.3ex]
    \bottomrule
    \end{tabularx}
   \caption{\small Snow \vspacesubtables}
   \label{tab:supp_CSC_r4}
\end{subtable}

\begin{subtable}[h]{\textwidth}
    \centering
    \def\arraystretch{\arraystretches}
    \setlength{\tabcolsep}{5pt}
    \begin{tabularx}{0.85\textwidth}{c|*{4}{Y}}
    \hline
    \toprule 
     Method & Source & TENT-continual~\cite{wang2021tent} & PETAL~\cite{Brahma2023probabilistic} & BESTTA (Ours) \\
     \hline
     Mean mIoU (\%) & \underline{41.4} & 36.3 & 26.7 & \textbf{45.0} \\
    \bottomrule
    \end{tabularx}
   \caption{\small Mean}
   \label{tab:supp_CSC_mean}
   \end{subtable}
}
\caption{Semantic segmentation results (mIoU in \%) on the Cityscapes-to-Cityscapes-C gradual test-time adaptation task. All experiments are evaluated using DeepLabV3Plus-ResNet50~\cite{chen2018deeplabv3+}. The \textbf{best} and \underline{second best} results are highlighted. \vspace{5em}}
\label{tab:supp_CSC}
\end{table*}

\topic{Further qualitative results}
Fig.~\ref{fig:supp_embeddings} illustrates the distributions of adapted embeddings of our method and the baselines in each channel. Our method effectively aligns the embedding distribution with the source distribution, irrespective of the weather conditions, demonstrating the effectiveness and stability of our method. On the contrary, the BN-based methods exhibit instability as their embedding distributions collapse to a single point. 
Fig.~\ref{fig:supp_Qual_SS} presents further predictions of our method and the baselines. Our method consistently outperforms the baselines across the weather conditions, with higher accuracy and lower noise.

{
\begin{figure*}[th!]
    \centering
    \includegraphics[width=0.95\linewidth]{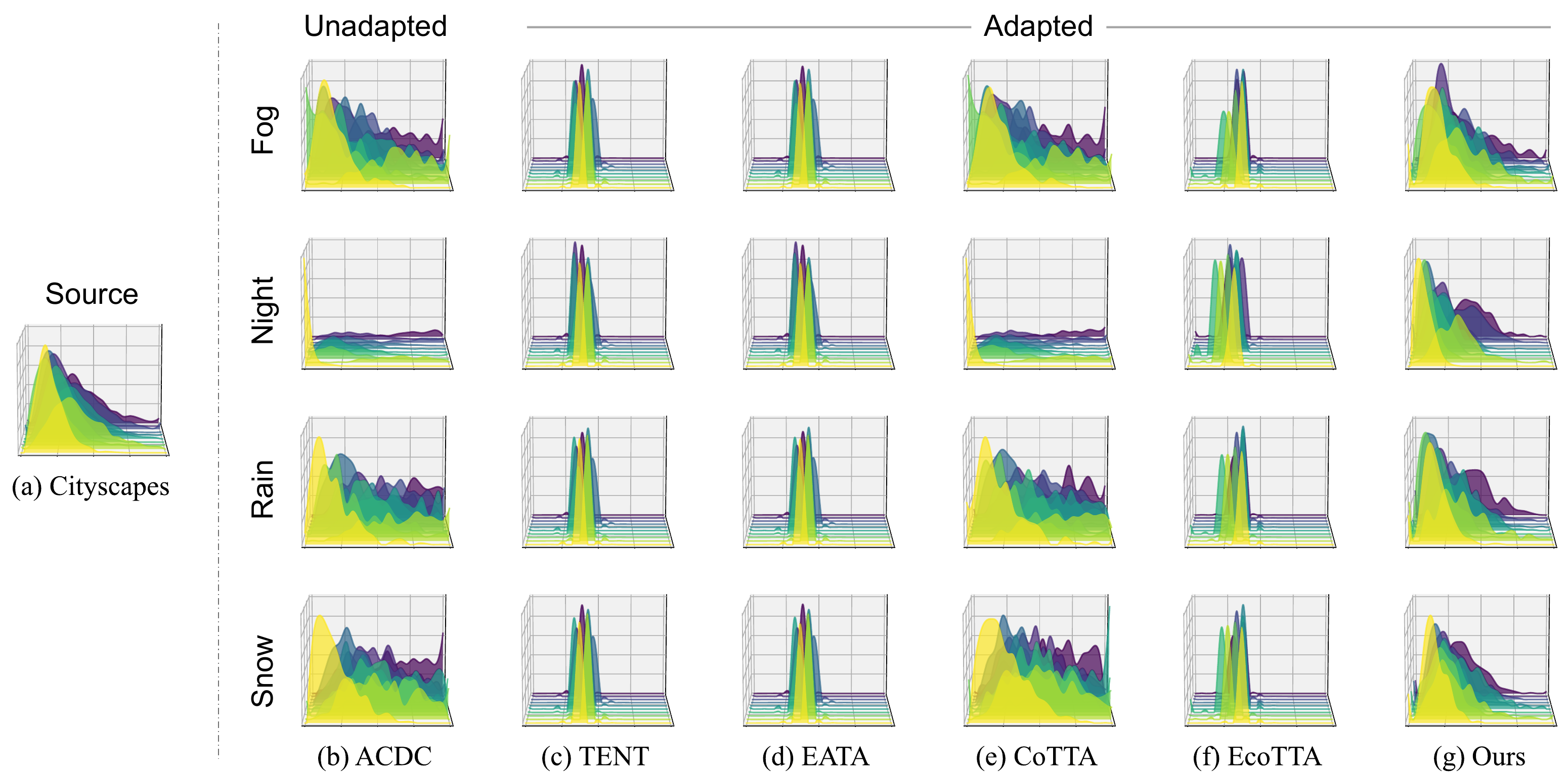}
    \caption{Embedding feature distributions of adapted embeddings in our method and the baselines. Our method successfully aligns the target distribution (ACDC) to the source distribution (Cityscapes). Our method successfully aligns the embedding distribution with the source distribution, whereas the BN-based methods (TENT~\cite{wang2021tent}, EATA~\cite{niu2022efficient}, EcoTTA~\cite{song2023ecotta}) exhibit instability as their embedding distributions collapse.    
    We use DeepLabV3Plus-ResNet50~\cite{chen2018deeplabv3+} pretrained on Cityscapes~\cite{cordts2016cityscapes} in this experiment.
    }
    \label{fig:supp_embeddings}
\end{figure*}

\begin{figure*}[th!]
    \centering
    \includegraphics[width=0.95\linewidth]{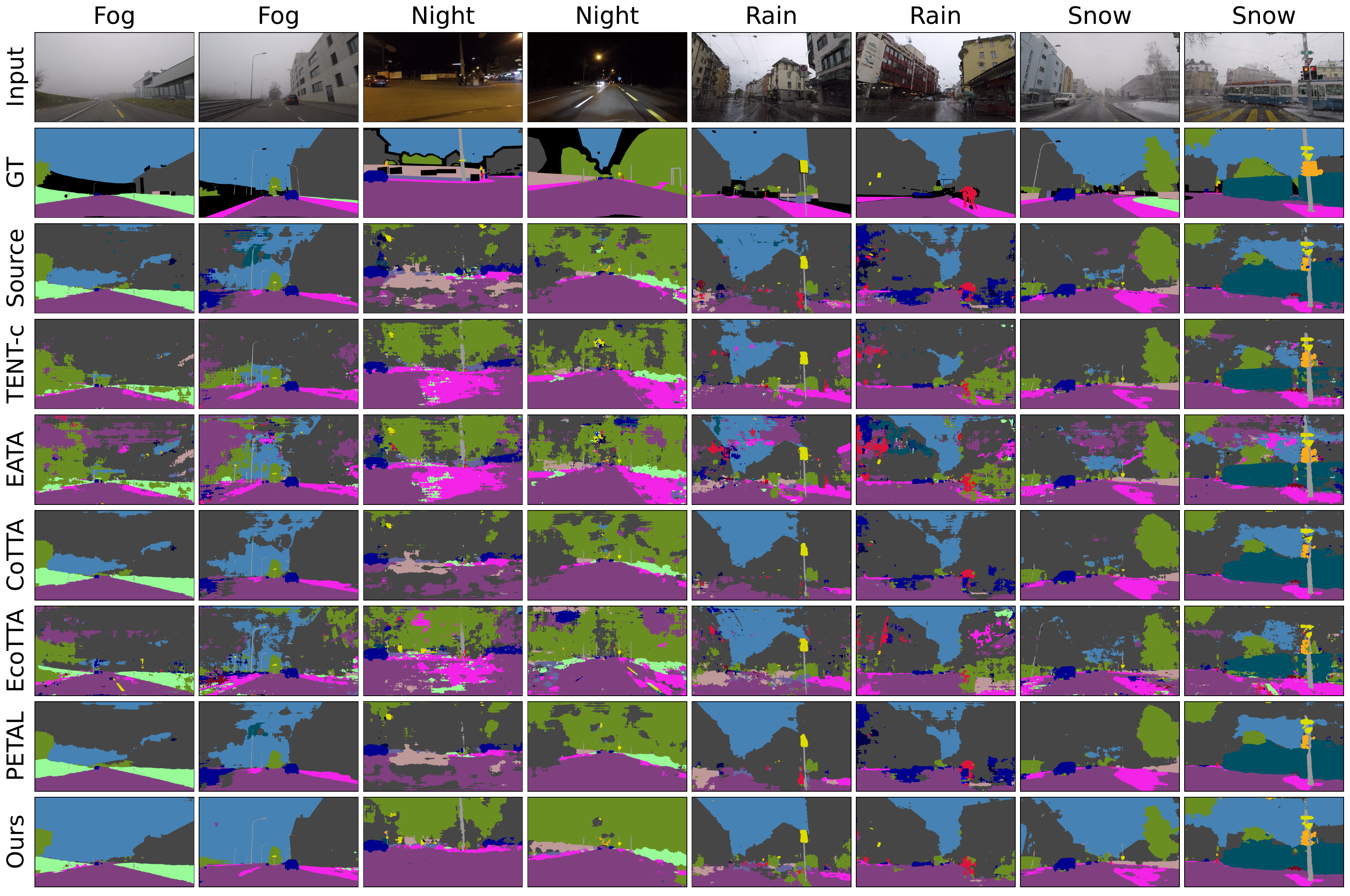}
    \caption{Further qualitative comparison of semantic segmentation on Cityscapes-to-ACDC single image continual test-time adaptation task. All experiments are evaluated using DeepLabV3Plus-ResNet50~\cite{chen2018deeplabv3+}. TENT-c denotes the TENT-continual~\cite{wang2021tent} method.
    }
    \label{fig:supp_Qual_SS}
\end{figure*}
}
}
}



\end{document}